\documentclass{article}
\usepackage{latexsym,amssymb,amsmath} 
\usepackage[colon]{natbib}
\usepackage{path}
\usepackage{url}
\usepackage{subcaption}
\usepackage{tikz}
\usetikzlibrary{arrows}
\usepackage{verbatim}
\usepackage{circuitikz}
\usetikzlibrary{positioning}
\usepackage{tkz-graph}
\usepackage{fancyhdr}
\usepackage{booktabs}
\usepackage{courier}
\usetikzlibrary{decorations.markings}
\usetikzlibrary{shapes,snakes}
\usepackage[toc,page]{appendix}
\usepackage{verbatim}
\usepackage{alltt}
\usepackage{sverb}
\usepackage{listings}
\usepackage{color}

\numberwithin{equation}{section}
\numberwithin{table}{section}
\numberwithin{figure}{section}

\definecolor{codegreen}{rgb}{0,0.6,0}
\definecolor{codegray}{rgb}{0.5,0.5,0.5}
\definecolor{codepurple}{rgb}{0.58,0,0.82}
\definecolor{backcolour}{rgb}{0.95,0.95,0.95}
\tikzstyle{block} = [draw,minimum size=2.5em, outer sep=2]

\begin{document}

\thispagestyle{fancy}

\tikzset{->-/.style={decoration={
  markings,
  mark=at position #1 with {\arrow{>}}},postaction={decorate}}}

\begin{center}
\textbf{\Large Recurrent Memory Array Structures} \\[24pt]
\textbf{Technical Report} \\[10pt]
 Kamil M Rocki\footnote{kmrocki@us.ibm.com}\\[2pt] \emph{IBM Research, San Jose, 95120, USA}  \\[6pt]
\end{center}

\vspace{12pt}


\begin{abstract}
The following report introduces ideas augmenting standard Long Short Term Memory (LSTM) architecture with multiple memory cells per hidden unit in order to improve its generalization capabilities. It considers both deterministic and stochastic variants of memory operation. It is shown that the nondeterministic Array-LSTM approach improves state-of-the-art performance on character level text prediction achieving 1.402 BPC\footnote{bits per character} on enwik8 dataset. Furthermore, this report estabilishes baseline neural-based results of 1.12 BPC and 1.19 BPC for enwik9 and enwik10 datasets respectively.

\end{abstract}
\vspace{12pt}

\section{Background}
It has been argued that the ability to compress arbitrary redundant patterns into short, compact representations may require an understanding that is equivalent to general artificial intelligence \citep{MacKay:itp, Hutter:04uaibook}. One example of such a process is demonstrated by learning to predict text a letter at a time. A strong connection between compression and prediction was shown \citep{shannon1951prediction}. Therefore, this report considers experiments on natural wikipedia text corpora, however the algorithms can be in principle applied to any sequences of patterns.

\section{Simple RNN}
\begin{figure}[h]
 \centering

\begin{tikzpicture}
    
    \node[] at (0,-3) (h_prev) {$\dots$};
    \node[block] at (1,-3) (h) {$h^{t-1}$};
    \node[block] at (1,-4) (x) {$x^{t}$};
    \node[block] at (9.9,-3) (h2) {$h^{t}$};
    \node[] at (11,-3) (h_next) {$\dots$};
    \node[draw,circle] at (6,-3) (tanh) {$tanh$};
    \draw[->, line width=0.6] (h_prev.east) -- (h);
    \draw[->, line width=0.6] (h.east) -- (tanh);
    \draw[->, line width=0.6] (tanh.east) -- (h2);
    \draw[->, line width=0.6] (h2.east) -- (h_next);

\node[label={[label distance=0.5cm,text depth=-1ex,rotate=0]right:internal state}] at (1.5,-3.2) {};

\node[label={[label distance=0.5cm,text depth=-1ex,rotate=0]right:$U$}] at (0.85,-2.65) {};
\node[label={[label distance=0.5cm,text depth=-1ex,rotate=0]right:$W$}] at (0.83,-3.7) {};
     \draw[->-=.94, line width=0.3] -- ([xshift=0.516cm, yshift=-0.05em]x) ++(0,0.0) -- ++(0.3,0) to [out=0,in=180] ++(0.45,1.028);

    \end{tikzpicture}

\caption{Simple RNN unit (omitted implementation specifics); $h^t$ - internal (hidden) state at time step $t$; $x$ are inputs, $y$ are optional outputs to be emitted; All connections are learnable parameters. No explicit asynchronous memory, implicit history aggregation only through hidden states $h$. Omitted bias terms for brevity.}
\label{fig:srnn}
\end{figure}
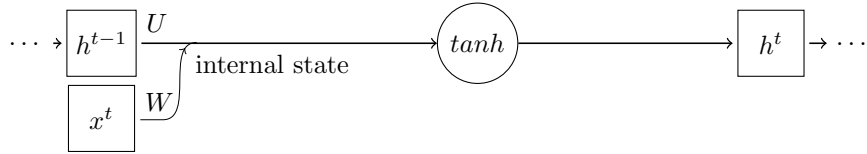

 A standard recurrent neural network (so called simple RNN, cite) is composed of a matrix of connections between its inputs and hidden states $W$, and a matrix $U$, connecting hidden states in consecutive time steps. In such a simple RNN architecture the entire history of observations in aggregated in hidden states of neurons (fig \ref{fig:srnn}). States ($h^t$) are determined by previous states ($h^{t-1}$) and feedforward immediate inputs ($x^{t}$). This architecture is deterministic, for 2 identical $h^{t-1}$, $x^t$ there will be 2 identical outputs $h^{t}$. A single time step update can be expressed with an equation (1) or equivalently using the following graphical representation as shown in fig. \ref{fig:srnn}.

 \begin{equation}
h^t = tanh({W x^t + U h^{t-1} + b})
\end{equation}

\section{Memory structures}
A simple RNN architecture does not handle long-range interactions and multiple simultaneous context well \citep{Bengio-trnn94}. However, it can be modified in order to make learning long-term dependencies easier. One solution to the problem is to change the way of interactions between hidden units, i.e. add multiplicative connections \citep{ICML2011Sutskever_524}. Another is by adding explicit memory cells. Networks involving register-like functionality allowing hidden states to store and load its contents in an asynchronous way have been very successful recently. Examples of such architectures include Long-Short Term Memory \citep{Hochreiter:1997:LSM:1246443.1246450}{} and Gated Recurrent Unit (GRU) \citep{Chung-et-al-TR2014} networks.

\subsection{LSTM}
 This section describes a standard LSTM network used in our experiments and serving as a foundation for array extensions. Equations 3.1-3.6 define a single LSTM time step update.
 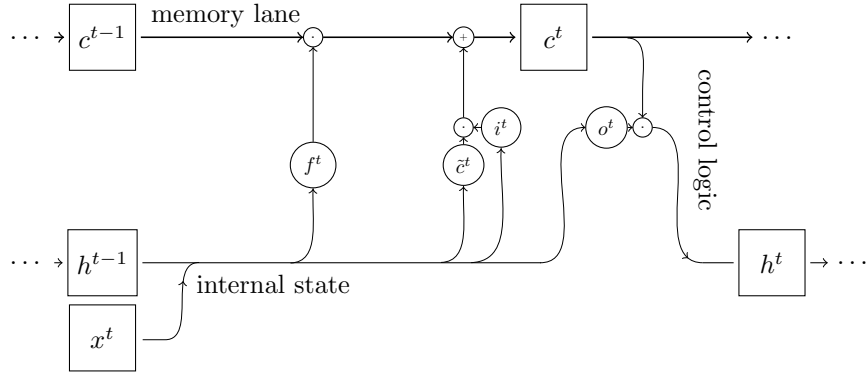
\begin{figure}[h]
 \centering
 \begin{tikzpicture}
    \foreach \y [evaluate = \y as \expr using (\y*0.4-1.2), evaluate = \y as \exprr using (\y-3.5), evaluate = \y as \exprrr using  int(3-\y)]  in {0} {
    	   \node[] at (0,-\y) (input_prev\y) {$\dots$};
        \node[block] at (1,-\y) (input\y) {$c^{t-1}$};
        \node[draw,circle, xscale=0.6, yscale=0.6]at (3.8,-\y) (f_\y) {.};
         \node[draw,circle, xscale=0.4, yscale=0.4]at (5.8,-\y) (f2_\y) {+};
	 \node[draw,circle, xscale=0.75, yscale=0.75] at (-\expr+2.6,-1.7) (f_block\y) {$f^{t}$};
   \node[draw,circle, xscale=0.75, yscale=0.75] at (-\expr+4.6,-1.7) (i_block\y) {$\tilde{c}^{t}$};
   \node[draw,circle, xscale=0.75, yscale=0.75] at (6.3,\expr) (g_block\y) {$i^{t}$};
   \node[draw,circle, xscale=0.75, yscale=0.75] at (7.7,\expr) (o_block\y) { $o^{t}$};
   \node[draw,circle, xscale=0.6, yscale=0.6] at (-\expr+4.6,\expr) (ig_block\y) {.};
   \node[draw,circle, xscale=0.6, yscale=0.6] at (\expr+9.38,\expr) (oc_block\y) {.};
       
        \node[block] at (7,-\y) (block\y) {$c^t$};
        \node[] at (10,-\y) (input_next\y) {$\dots$};
        \draw[->, line width=0.6] (input_prev\y.east) -- (input\y);
        \draw[->, line width=0.6] (input\y) --  (f_\y.west);
	\draw[->, line width=0.6] (f_\y.east) --(f2_\y.west);
	\draw[->, line width=0.6] (f2_\y.east) --(block\y);
        \draw[->, line width=0.6] (block\y.east) -- (input_next\y);
        
    }
    
    \node[] at (0,-3) (h_prev) {$\dots$};
    \node[block] at (1,-3) (h) {$h^{t-1}$};
    \node[block] at (1,-4) (x) {$x^{t}$};
    \node[block] at (9.9,-3) (h2) {$h^{t}$};
    \node[] at (11,-3) (h_next) {$\dots$};

    \draw[->] (h_prev.east) -- (h);
    \draw[->] (h2.east) -- (h_next);
    \draw[] (h.east) ++(5.3,0) -- (h);
    \draw[] (h.east) ++(7.45,0) -- (h2);


\node[label={[label distance=0.5cm,text depth=-1ex,rotate=0]right:memory lane}] at (0.9,0.4) {};
\node[label={[label distance=0.5cm,text depth=-1ex,rotate=-90]right:control logic}] at (9.0,0.2) {};
\node[label={[label distance=0.5cm,text depth=-1ex,rotate=0]right:internal state}] at (1.5,-3.2) {};



    \draw[->-=.99, line width=0.3] (f_block0.south) ++(-0.3,-0.995) to [out=2,in=-93] (f_block0.south);

	\draw[->, line width=0.3] (f_block0.north)  to [out=90,in=-90] (f_0);
	\draw[->, line width=0.3] (ig_block0.north)  to [out=90,in=-90] (f2_0);
    \draw[->-=.99, line width=0.3] (i_block0.south) ++(-0.3,-1.02) to [out=2,in=-93] (i_block0.south);

    \draw[->-=.99, line width=0.3] (g_block0.south) ++(-0.4,-1.531) to [out=2,in=-93] (g_block0.south);

	\draw[<-, line width=0.3] (o_block0.west)  to [out=180,in=0] ++(-0.6,-1.8);
    \draw[<-, line width=0.3] (oc_block0) -- ++(-0,0.49) to [out=90,in=0] ++(-0.2,0.7);

    \draw[->, line width=0.3] (i_block0.north) -- (ig_block0.south);

    \draw[->, line width=0.3] (g_block0.west) -- (ig_block0.east);

    \draw[->, line width=0.3] (o_block0.east) -- (oc_block0.west);

    \draw[->-=.94, line width=0.3] -- ([xshift=-4.266cm, yshift=-11.45em]f2_0) ++(0,0.0) -- ++(0.3,0) to [out=0,in=180] ++(0.45,1.028);

    \draw[->-=.95, line width=0.3] -- (oc_block0) to [out=-0,in=180] ++(0.8,-1.8);

    \end{tikzpicture}
    \caption{LSTM, biases and nonlinearites omitted for brevity; The memory content $c^{t}$ is set according to the gates' activations which are in turn driven by the bottom-up input $x^t$ and previous internal state $h^{t-1}$.}
    \label{fig:LSTM}
\end{figure}


\begin{equation}
f^t = \sigma({W_f x^t + U_f h^{t-1} + b_f})
\end{equation}
\begin{equation}
i^t = \sigma({W_i x^t + U_i h^{t-1} + b_i})
\end{equation}
\begin{equation}
o^t = \sigma({W_o x^t + U_o h^{t-1} + b_o})
\end{equation}
\begin{equation}
\tilde{c}^{t} = tanh({W_c x^t + U_c h^{t-1} + b_c})
\end{equation}
\begin{equation}
c^{t} = f_t \odot c^{t-1} + i_t \odot \tilde{c}^{t}
\end{equation}
\begin{equation}
h^{t} = o_t \odot tanh(c^{t})
\end{equation}

\subsection{State-sharing Memory: Array-LSTM}
\begin{figure*}[h]
    \centering
    \begin{subfigure}[t]{0.5\textwidth}
        \centering
        \includegraphics[height=2.5in]{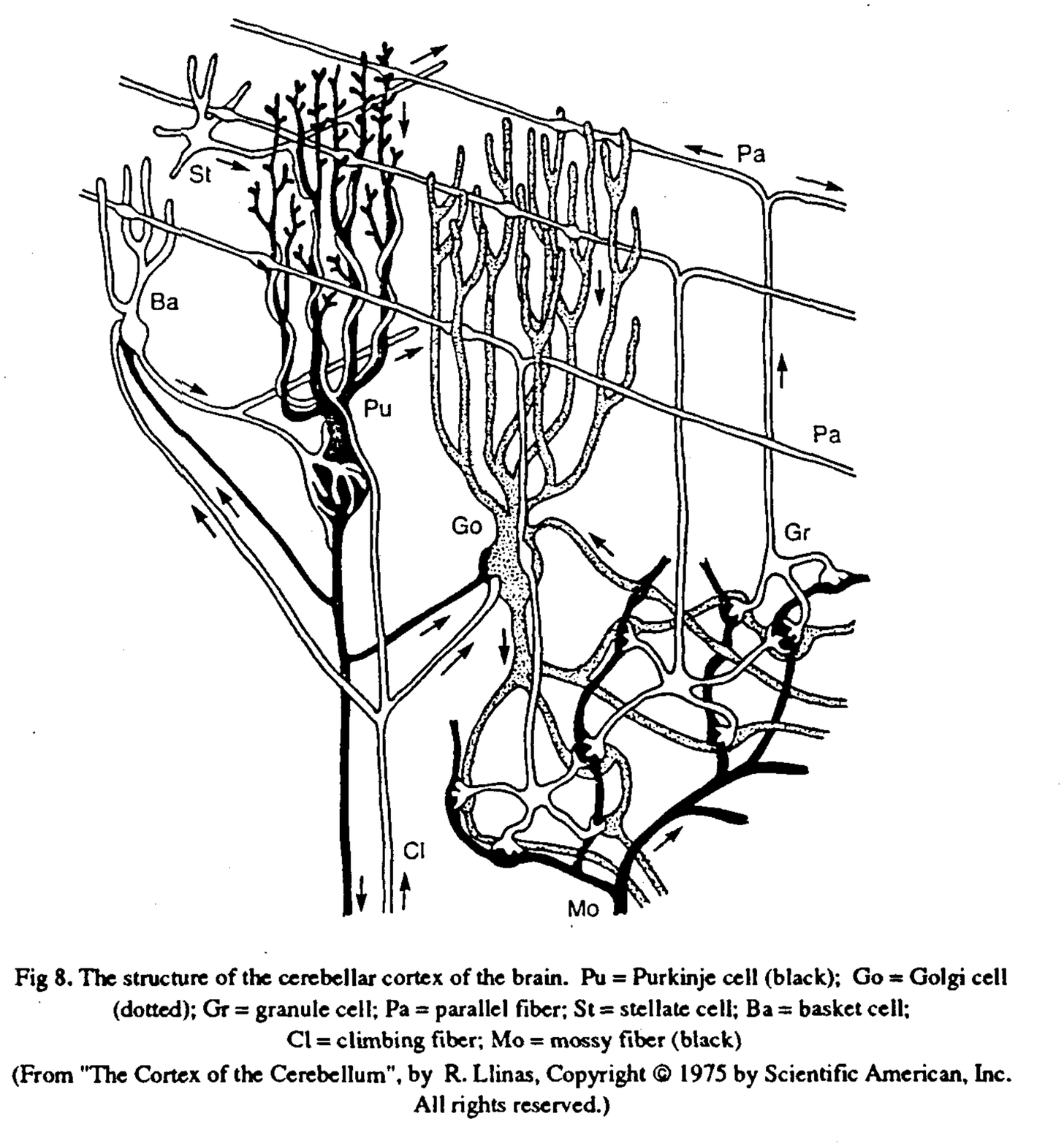}
    \end{subfigure}%
    ~ 
    \begin{subfigure}[t]{0.5\textwidth}
        \centering
        \includegraphics[height=2.5in]{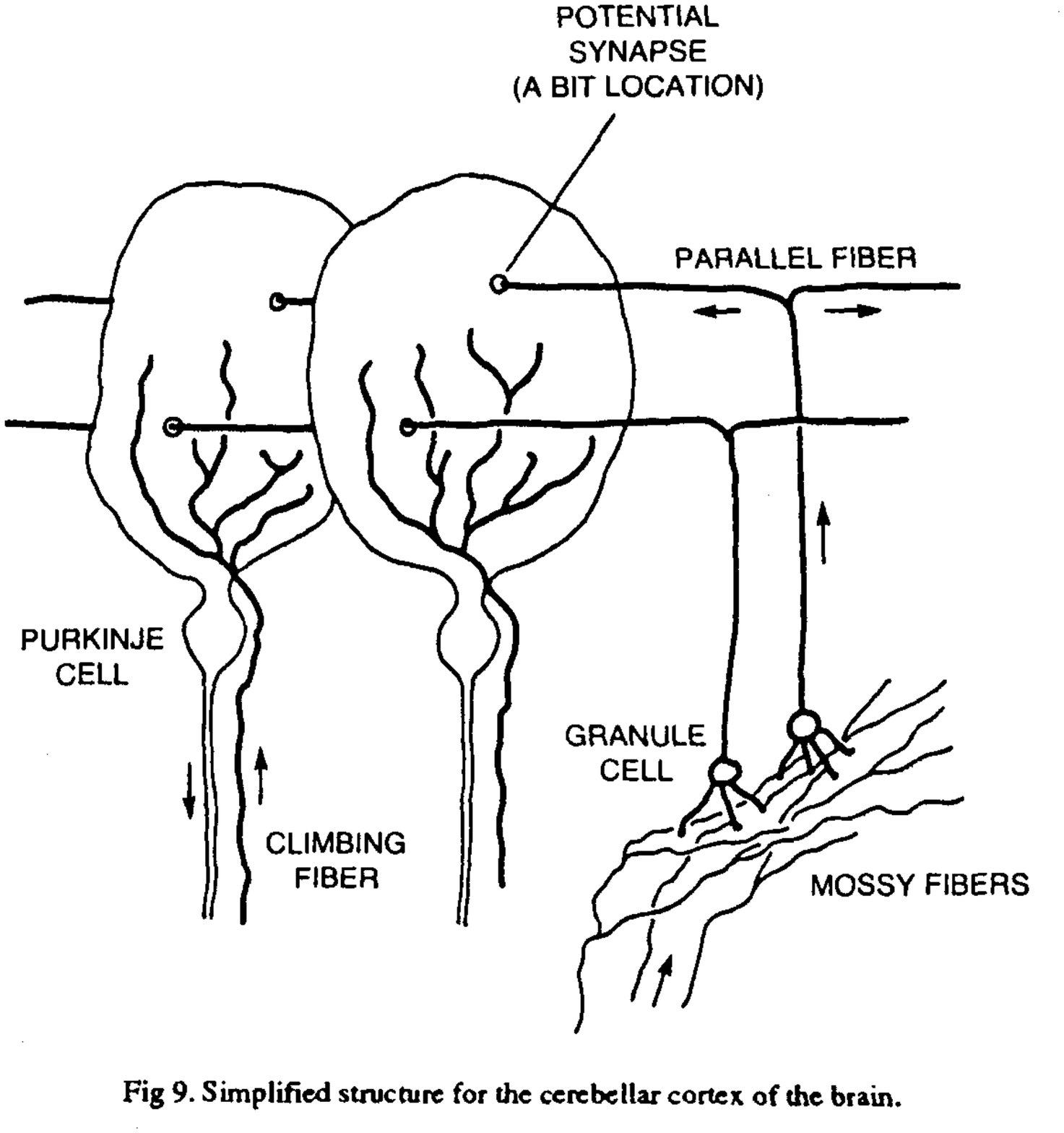}
    \end{subfigure}
    \caption{The structure of the cerebellar cortex \citep{Kanerva:1988:SDM:534853}}
\label{fig:cereb}
\end{figure*}

 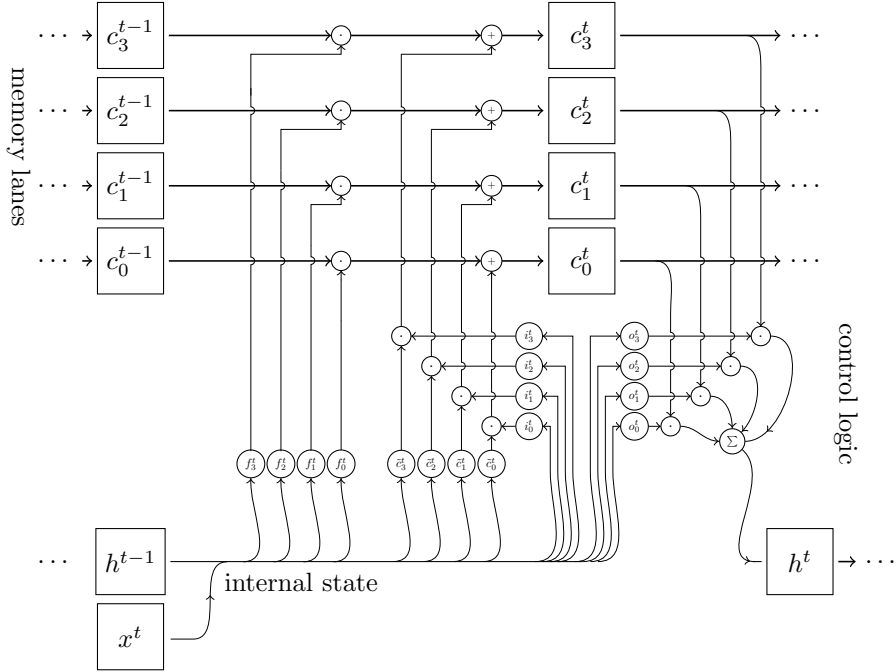
\begin{figure}[h]
 \centering
 \begin{tikzpicture}
    \foreach \y [evaluate = \y as \expr using (\y*0.4-1.2), evaluate = \y as \exprr using (\y-3.5), evaluate = \y as \exprrr using  int(3-\y)]  in {0,1,2,3} {
    	   \node[] at (0,-\y) (input_prev\y) {$\dots$};
        \node[block] at (1,-\y) (input\y) {$c_{\exprrr}^{t-1}$};
        \node[draw,circle, xscale=0.6, yscale=0.6]at (3.8,-\y) (f_\y) {.};
         \node[draw,circle, xscale=0.4, yscale=0.4]at (5.8,-\y) (f2_\y) {+};
	 \node[draw,circle, xscale=0.45, yscale=0.45] at (-\expr+2.6,-5.7) (f_block\y) {$f_\y^{t}$};
   \node[draw,circle, xscale=0.45, yscale=0.45] at (-\expr+4.6,-5.7) (i_block\y) {$\tilde{c}_\y^{t}$};
   \node[draw,circle, xscale=0.45, yscale=0.45] at (6.3,\expr-4) (g_block\y) {$i_\y^{t}$};
   \node[draw,circle, xscale=0.45, yscale=0.45] at (7.7,\expr-4) (o_block\y) {$o_\y^{t}$};
   \node[draw,circle, xscale=0.6, yscale=0.6] at (-\expr+4.6,\expr-4) (ig_block\y) {.};
   \node[draw,circle, xscale=0.6, yscale=0.6] at (\expr+9.38,\expr-4) (oc_block\y) {.};
       
        \draw[<-, line width=0.3] (f_\y) -- ++(0,-.25) -- ++(\expr,0) -- ++(0,-.55);
        \draw[<-, line width=0.3 ] (f2_\y) -- ++(0,-.25) -- ++(\expr,0);
        \node[block] at (7,-\y) (block\y) {$c_{\exprrr}^t$};
        \node[] at (10,-\y) (input_next\y) {$\dots$};
        \draw[->, line width=0.6] (input_prev\y.east) -- (input\y);
        \draw[->, line width=0.6] (input\y) --  (f_\y.west);
	\draw[->, line width=0.6] (f_\y.east) --(f2_\y.west);
	\draw[->, line width=0.6] (f2_\y.east) --(block\y);
        \draw[->, line width=0.6] (block\y.east) -- (input_next\y);
        
    }
    
    \node[] at (0,-7) (h_prev) {$\dots$};
    \node[block] at (1,-7) (h) {$h^{t-1}$};
    \node[block] at (1,-8) (x) {$x^{t}$};
    \node[block] at (9.9,-7) (h2) {$h^{t}$};
    \node[] at (11,-7) (h_next) {$\dots$};

    \draw[] (h.east) ++(5.48,0) -- (h);
    \draw[] (h.east) ++(7.75,0) -- (h2);
    \draw[->, line width=0.6] (h2.east) -- (h_next);

    \node[draw,circle, xscale=0.4, yscale=0.4]at (9,-5.4) (sum) {$ \sum $};

\node[label={[label distance=0.5cm,text depth=-1ex,rotate=-90]right:memory lanes}] at (-0.5,0.2) {};
\node[label={[label distance=0.5cm,text depth=-1ex,rotate=-90]right:control logic}] at (10.5,-3.2) {};
\node[label={[label distance=0.5cm,text depth=-1ex,rotate=0]right:internal state}] at (1.5,-7.2) {};

    \draw[line width=0.3]  -- ([xshift=-1.2cm, yshift=-1.85em]f_0) ++(0,-.05) -- ++(0,-.25) arc(90:-90:0.05cm) -- ++(0,-.75) -- ++(0,-.15) arc(90:-90:0.05cm) -- ++(0,-.75) -- ++(0,-.15) arc(90:-90:0.05cm) -- ++(0,-2.48);
    \draw[line width=0.3]  -- ([xshift=-0.8cm, yshift=-2.28em]f_0)  ++(0,-.25) ++(0,-.75) -- ++(0,-.15) arc(90:-90:0.05cm) -- ++(0,-.75) -- ++(0,-.15) arc(90:-90:0.05cm) -- ++(0,-2.48);
    \draw[line width=0.3]  -- ([xshift=-0.4cm, yshift=-5.82em]f_0) ++(0,-.75) -- ++(0,-.15) arc(90:-90:0.05cm) -- ++(0,-2.48);
    \draw[line width=0.3]  -- ([xshift=-0.0cm, yshift=-5.82em]f_0) ++(0,-1.75) -- ++(0,-1.72);

    \draw[line width=0.3] -- ([xshift=-1.2cm, yshift=-1.85em]f2_0) ++(0,0.405) -- ++(0,-.705) arc(90:-90:0.05cm) -- ++(0,-.75) -- ++(0,-.15) arc(90:-90:0.05cm) -- ++(0,-.75) -- ++(0,-.15) arc(90:-90:0.05cm) -- ++(0,-.82);
    \draw[line width=0.3]  -- ([xshift=-0.8cm, yshift=-2.28em]f2_0)  ++(0,.305) ++(0,-.75) -- ++(0,-.705) arc(90:-90:0.05cm) -- ++(0,-.75) -- ++(0,-.15) arc(90:-90:0.05cm) -- ++(0,-.9) arc(90:-90:0.05cm) -- ++(0,-.22);
    \draw[line width=0.3]  -- ([xshift=-0.4cm, yshift=-5.82em]f2_0) ++(0,-.2025) -- ++(0,-0.7) arc(90:-90:0.05cm) -- ++(0,-.9) arc(90:-90:0.05cm) -- ++(0,-.3) arc(90:-90:0.05cm) -- ++(0,-.22);
    \draw[line width=0.3]  -- ([xshift=-0.0cm, yshift=-5.82em]f2_0) ++(0,-1.1) -- ++(0,-.8) arc(90:-90:0.05cm) -- ++(0,-.3) arc(90:-90:0.05cm) -- ++(0,-.3) arc(90:-90:0.05cm) -- ++(0,-.22);

    \draw[->-=.99, line width=0.3] (f_block0.south) ++(-0.1,-1.115) to [out=2,in=-93] (f_block0.south);
    \draw[->-=.99, line width=0.3] (f_block1.south) ++(-0.1,-1.115) to [out=2,in=-93] (f_block1.south);
    \draw[->-=.99, line width=0.3] (f_block2.south) ++(-0.1,-1.115) to [out=2,in=-93] (f_block2.south);
    \draw[->-=.99, line width=0.3] (f_block3.south) ++(-0.1,-1.115) to [out=2,in=-93] (f_block3.south);

    \draw[->-=.99, line width=0.3] (i_block0.south) ++(-0.1,-1.115) to [out=2,in=-93] (i_block0.south);
    \draw[->-=.99, line width=0.3] (i_block1.south) ++(-0.1,-1.115) to [out=2,in=-93] (i_block1.south);
    \draw[->-=.99, line width=0.3] (i_block2.south) ++(-0.1,-1.115) to [out=2,in=-93] (i_block2.south);
    \draw[->-=.99, line width=0.3] (i_block3.south) ++(-0.1,-1.115) to [out=2,in=-93] (i_block3.south);

    \draw[<-, line width=0.3] (g_block0.east) ++(-0.0,0) -- ++(0.1,0) -- ++(0,-0.6) to [out=-90,in=0] ++(-0.2,-1.2);
    \draw[<-, line width=0.3] (g_block1.east) ++(-0.0,0) -- ++(0.2,0) -- ++(0,-1.0) to [out=-90,in=0] ++(-0.2,-1.2);
    \draw[<-, line width=0.3] (g_block2.east) ++(-0.0,0) -- ++(0.3,0) -- ++(0,-1.4) to [out=-90,in=0] ++(-0.2,-1.2);
    \draw[<-, line width=0.3] (g_block3.east) ++(-0.0,0) -- ++(0.4,0) -- ++(0,-1.8) to [out=-90,in=0] ++(-0.2,-1.2);

    \draw[<-, line width=0.3] (o_block0.west) ++(-0.0,0) -- ++(-0.1,0) -- ++(0,-0.6) to [out=-90,in=0] ++(-0.2,-1.2);
    \draw[<-, line width=0.3] (o_block1.west) ++(-0.0,0) -- ++(-0.2,0) -- ++(0,-1.0) to [out=-90,in=0] ++(-0.2,-1.2);
    \draw[<-, line width=0.3] (o_block2.west) ++(-0.0,0) -- ++(-0.3,0) -- ++(0,-1.4) to [out=-90,in=0] ++(-0.2,-1.2);
    \draw[<-, line width=0.3] (o_block3.west) ++(-0.0,0) -- ++(-0.4,0) -- ++(0,-1.8) to [out=-90,in=0] ++(-0.2,-1.2);

    \draw[<-, line width=0.3] (oc_block0) -- ++(-0,0.35) arc(-90:90:0.05cm) -- ++(-0,0.3) arc(-90:90:0.05cm) -- ++(-0,0.3) arc(-90:90:0.05cm) -- ++(-0,0.25) to [out=90,in=0] ++(-0.2,0.7);
    \draw[<-, line width=0.3] (oc_block1) -- ++(-0,0.35) arc(-90:90:0.05cm) -- ++(-0,0.3) arc(-90:90:0.05cm) -- ++(-0,0.9) arc(-90:90:0.05cm) -- ++(0,0.45) to [out=90,in=0] ++(-0.2,0.5);
    \draw[<-, line width=0.3] (oc_block2) -- ++(-0,0.35) arc(-90:90:0.05cm) -- ++(-0,0.9) arc(-90:90:0.05cm) -- ++(0,0.9) arc(-90:90:0.05cm) -- ++(0,0.45) to [out=90,in=0] ++(-0.2,0.5);
    \draw[<-, line width=0.3] (oc_block3) -- ++(-0,0.95) arc(-90:90:0.05cm) -- ++(0,0.9) arc(-90:90:0.05cm) -- ++(0,0.9) arc(-90:90:0.05cm) -- ++(0,0.45) to [out=90,in=0] ++(-0.2,0.5);

    \draw[->, line width=0.3] (i_block0.north) -- (ig_block0.south);
    \draw[->, line width=0.3] (i_block1.north) -- (ig_block1.south);
    \draw[->, line width=0.3] (i_block2.north) -- (ig_block2.south);
    \draw[->, line width=0.3] (i_block3.north) -- (ig_block3.south);

    \draw[->, line width=0.3] (g_block0.west) -- (ig_block0.east);
    \draw[->, line width=0.3] (g_block1.west) -- (ig_block1.east);
    \draw[->, line width=0.3] (g_block2.west) -- (ig_block2.east);
    \draw[->, line width=0.3] (g_block3.west) -- (ig_block3.east);

    \draw[->, line width=0.3] (o_block0.east) -- (oc_block0.west);
    \draw[->, line width=0.3] (o_block1.east) -- (oc_block1.west);
    \draw[->, line width=0.3] (o_block2.east) -- (oc_block2.west);
    \draw[->, line width=0.3] (o_block3.east) -- (oc_block3.west);

    \draw[->-=.94, line width=0.3] -- ([xshift=-4.266cm, yshift=-22.85em]f2_0) ++(0,0.0) -- ++(0.3,0) to [out=0,in=180] ++(0.45,1.028);

    \draw[->-=.90, line width=0.3] -- (oc_block3) to [out=0,in=0] (sum);
    \draw[->-=.99, line width=0.3] -- (oc_block2) to [out=0,in=45] (sum);
    \draw[->-=.99, line width=0.3] -- (oc_block1) to [out=0,in=90] (sum);
    \draw[->-=.99, line width=0.3] -- (oc_block0) to [out=0,in=180] (sum);
    \draw[->-=.98, line width=0.3] -- (sum) to [out=-40,in=180] ++(0.3,-1.6);

    \end{tikzpicture}
    \caption{Array-LSTM with 4 memory cells per hidden control unit; modulated vs modulating connections, omitted nonlinearities and biases for brevity; It becomes a standard LSTM when only 1 memory cell per hidden unit is present}
    \label{fig:ALSTM}
\end{figure}

Cerebellar cortex structure as described in \citep{Kanerva:1988:SDM:534853} among others: Fig. \ref{fig:cereb} shows array-like structure in what seems to be Random-Access Memory in cerebellum. It has served as an inspiration for the Array-LSTM architecture. The main idea is that instead of building hierarchies of layers (as in stacked LSTM \citep{journals/corr/Graves13}, Gated-Feedback RNN \citep{DBLP:journals/corr/ChungGCB15}) keep a single layer, but build more complex memory structures inside a RNN unit (a similar line of thinking has been explored by contructing a more complex transition function inside a layer \citep{Pascanu+et+al-ICLR2014}). We want to create a $bottleneck$ by sharing internal states, forcing the learning procedure to $pool$ similar or interchangeable content using memory cells belonging to one hidden unit (analogy would be that a word related to car would activate a particular hidden state and each memory cell could be interchangeably used if it represents a particular car type, therefore externally the choice of a particular memory cell would not be relevant. Similar concepts exist already in convnets, i.e. spatial pooling, the hidden state should work as a complex cell pooling multiple possible substates. Figure \ref{fig:ALSTM} and equations 3.1-3.6 describe a single Array-LSTM architecture. Note the similarity between figures \ref{fig:cereb} and \ref{fig:ALSTM}. Furthermore, We consider modifications of the presented simple Array-LSTM architecture which are meant to improve capacity, memory efficiency, learning time or generalization. This report shows two types of changes, one pertains to deterministic family of architectures and the other one to stochastic operations (working in a dropout-like fashion).

\label{arraysection}
\begin{equation}
f_k^t = \sigma({W_{fk} x^t + U_{fk} h^{t-1} + b_{fk}})
\end{equation}
\begin{equation}
i_k^t = \sigma({W_{ik} x^t + U_{ik} h^{t-1} + b_{ik}})
\end{equation}
\begin{equation}
o_k^t = \sigma({W_{ok} x^t + U_{ok} h^{t-1} + b_{ok}})
\end{equation}
\begin{equation}
\tilde{c}_k^{t} = tanh({W_{ck} x^t + U_{ck} h^{t-1} + b_{ck}})
\end{equation}
\begin{equation}
c_k^{t} = f_k^t \odot c_k^{t-1} + i_k^t \odot \tilde{c}_k^{t}
\end{equation}

\begin{equation}
h^{t} = \sum_{k}{o_k^t} \odot tanh(c_k^{t})
\end{equation}

\section{Deterministic Array-LSTM extensions}

\subsection{Lane selection: Soft attention}
\label{softatt}
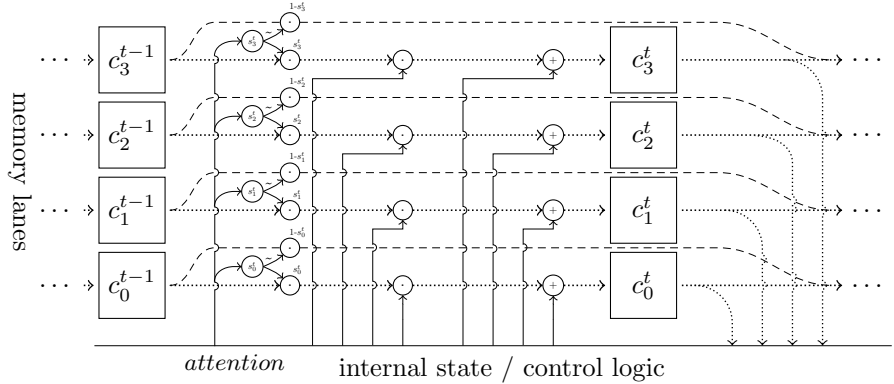
\begin{figure}
 \centering
 \begin{tikzpicture}
    \foreach \y [evaluate = \y as \expr using (\y*0.4-1.2), evaluate = \y as \exprr using (\y-3.5), evaluate = \y as \exprrr using  int(3-\y)]  in {0,1,2,3} {
           \node[] at (-0.8,-\y) (input_prev\y) {$\dots$};
        \node[block] at (0.2,-\y) (input\y) {$c_{\exprrr}^{t-1}$};
        \node[draw,circle, xscale=0.6, yscale=0.6]at (2.3,-\y + 0.5) (s0_\y) {.};
         \node[draw,circle, xscale=0.6, yscale=0.6]at (2.3,-\y) (s1_\y) {.};
        \node[draw,circle, xscale=0.6, yscale=0.6]at (3.8,-\y) (f_\y) {.};
         \node[draw,circle, xscale=0.4, yscale=0.4]at (5.8,-\y) (f2_\y) {+};
          \node[draw,circle, xscale=0.35, yscale=0.35] at (1.8,-\y + 0.25) (s_block\y) {$s_\exprrr^{t}$};
           \node[circle, xscale=0.35, yscale=0.35] at (2.4,-\y + 0.71) (s_lab1\y) {1-$s_\exprrr^{t}$};
           \node[circle, xscale=0.35, yscale=0.35] at (2.4,-\y + 0.21) (s_lab2\y) {$s_\exprrr^{t}$};
           \node[circle, xscale=0.7, yscale=0.7] at (2.02,-\y + 0.29) (s_lab2\y) {$\tilde{}$};
        \draw[<-, line width=0.3] (f_\y) -- ++(0,-.25) -- ++(\expr,0) -- ++(0,-.21);
        \draw[<-, line width=0.3 ] (f2_\y) -- ++(0,-.25) -- ++(\expr,0) -- ++(0,-.21);
         \draw[<-, line width=0.3 ] (s_block\y.west) -- ++(-0.1, 0) to [out=180,in=90] ++(-0.26,-0.2);

         \draw[->, line width=0.3 ] (s_block\y.east) to [out=-25,in=135] (s1_\y);
         \draw[->, line width=0.3 ] (s_block\y.east) to [out=25,in=-135] (s0_\y);
        \node[block] at (7,-\y) (block\y) {$c_{\exprrr}^t$};
        \node[] at (10,-\y) (input_next\y) {$\dots$};
        \draw[densely dotted, ->, line width=0.6] (input_prev\y.east) -- (input\y);
        \draw[densely dotted, ->, line width=0.6] (input\y) --  (s1_\y.west);
        \draw[densely dotted, ->, line width=0.6] (s1_\y.east) --(f_\y.west);
    \draw[densely dotted, ->, line width=0.6] (f_\y.east) --(f2_\y.west);
    \draw[densely dotted, ->, line width=0.6] (f2_\y.east) --(block\y);
        \draw[densely dotted, ->, line width=0.6] (block\y.east) -- (input_next\y);
        
    }

\draw[solid, line width=0.2]  -- ([xshift=-5cm, yshift=-10.82em]f2_0) ++(-1.1, 0) -- ++(10.6, 0);

\node[label={[label distance=0.5cm,text depth=-1ex,rotate=-90]right:memory lanes}] at (-1.3,0.2) {};
\node[label={[label distance=0.5cm,text depth=-1ex,rotate=0]right:internal state / control logic}] at (2.2,-4) {};
\node[label={[label distance=0.5cm,text depth=-1ex,rotate=0, font=\small]right:$attention$}] at (0.15,-3.95) {};

    \draw[line width=0.3]  -- ([xshift=-0.5cm, yshift=9.82em]s_block3)  ++(0,-.65)
    arc(90:-90:0.05cm) -- ++(0,-.15)
    -- ++(0,-.25) arc(90:-90:0.05cm) 
    -- ++(0,-.41) arc(90:-90:0.05cm)
    -- ++(0,-.39) arc(90:-90:0.05cm)
    -- ++(0,-.41) arc(90:-90:0.05cm)
    -- ++(0,-.39) arc(90:-90:0.05cm)
    -- ++(0,-.41) arc(90:-90:0.05cm)
    -- ++(0,-.75);

    \draw[line width=0.3]  -- ([xshift=-1.2cm, yshift=-1.85em]f_0) ++(0,.2)
    arc(90:-90:0.05cm) -- ++(0,-.15)
    -- ++(0,-.25) arc(90:-90:0.05cm) 
    -- ++(0,-.41) arc(90:-90:0.05cm)
    -- ++(0,-.39) arc(90:-90:0.05cm)
    -- ++(0,-.41) arc(90:-90:0.05cm)
    -- ++(0,-.39) arc(90:-90:0.05cm)
    -- ++(0,-.75);

    \draw[line width=0.3]  -- ([xshift=-0.8cm, yshift=-2.28em]f_0)  ++(0,-.65)
    arc(90:-90:0.05cm) -- ++(0,-.15)
    -- ++(0,-.25) arc(90:-90:0.05cm) 
    -- ++(0,-.41) arc(90:-90:0.05cm)
    -- ++(0,-.39) arc(90:-90:0.05cm)
    -- ++(0,-.75);

    \draw[line width=0.3]  -- ([xshift=-0.4cm, yshift=-5.82em]f_0) ++(0,-0.4)
    arc(90:-90:0.05cm) -- ++(0,-.15)
    -- ++(0,-.25) arc(90:-90:0.05cm) 
    -- ++(0,-0.75);

    \draw[line width=0.3]  -- ([xshift=-0.0cm, yshift=-5.82em]f_0) ++(0,-1.41) -- ++(0,-0.35);

    \draw[line width=0.3]  -- ([xshift=-1.2cm, yshift=-1.85em]f2_0) ++(0,.2)
    arc(90:-90:0.05cm) -- ++(0,-.15)
    -- ++(0,-.25) arc(90:-90:0.05cm) 
    -- ++(0,-.41) arc(90:-90:0.05cm)
    -- ++(0,-.39) arc(90:-90:0.05cm)
    -- ++(0,-.41) arc(90:-90:0.05cm)
    -- ++(0,-.39) arc(90:-90:0.05cm)
    -- ++(0,-.75);

    \draw[line width=0.3]  -- ([xshift=-0.8cm, yshift=-2.28em]f2_0)  ++(0,-.65)
    arc(90:-90:0.05cm) -- ++(0,-.15)
    -- ++(0,-.25) arc(90:-90:0.05cm) 
    -- ++(0,-.41) arc(90:-90:0.05cm)
    -- ++(0,-.39) arc(90:-90:0.05cm)
    -- ++(0,-.75);

    \draw[line width=0.3]  -- ([xshift=-0.4cm, yshift=-5.82em]f2_0) ++(0,-0.4)
    arc(90:-90:0.05cm) -- ++(0,-.15)
    -- ++(0,-.25) arc(90:-90:0.05cm) 
    -- ++(0,-0.75);

    \draw[line width=0.3]  -- ([xshift=-0.0cm, yshift=-5.82em]f2_0) ++(0,-1.41) -- ++(0,-0.35);

    \draw[densely dotted, <-, line width=0.5] (oc_block0) ++(-0,1.4) -- ++(-0,0.3) to [out=90,in=0] ++(-0.5,0.5);

    \draw[densely dotted, <-, line width=0.5] (oc_block1) ++(-0,1.0) -- ++(-0,0.75) arc(-90:90:0.05cm) -- ++(-0,0.25) arc(-90:90:0.05cm) -- ++(0,0.1) to [out=90,in=0] ++(-0.5,0.5);

    \draw[densely dotted, <-, line width=0.5] (oc_block2) ++(-0,0.6) -- ++(-0,0.75) arc(-90:90:0.05cm) -- ++(-0,0.02) arc(-90:90:0.05cm) -- ++(-0,0.79) arc(-90:90:0.05cm) -- ++(-0,0.02) arc(-90:90:0.05cm) -- ++(0,0.45) to [out=90,in=0] ++(-0.5,0.36);

    \draw[densely dotted, <-, line width=0.5] (oc_block3) ++(-0,0.2) -- ++(-0,0.75) arc(-90:90:0.05cm) -- ++(0,0.9) arc(-90:90:0.05cm) -- ++(0,0.9) arc(-90:90:0.05cm) -- ++(0,0.45) to [out=90,in=0] ++(-0.5,0.5);

    \draw[densely dashed, line width=0.3] (input0) ++(0.5,0) to [out=0,in=180] ++(0.6,0.5) -- (s0_0) -- ++(5.7,0) to [out=0,in=180] ++(1.5,-0.5);

    \draw[densely dashed, line width=0.3] (input1) ++(0.5,0) to [out=0,in=180] ++(0.6,0.5) -- (s0_1) -- ++(5.7,0) to [out=0,in=180] ++(1.5,-0.5);
    \draw[densely dashed, line width=0.3] (input2) ++(0.5,0) to [out=0,in=180] ++(0.6,0.5) -- (s0_2) -- ++(5.7,0) to [out=0,in=180] ++(1.5,-0.5);
    \draw[densely dashed, line width=0.3] (input3) ++(0.5,0) to [out=0,in=180] ++(0.6,0.5) -- (s0_3) -- ++(5.7,0) to [out=0,in=180] ++(1.5,-0.5);

    \end{tikzpicture}
    \caption{Lane selection through soft attention, solid lines are control logic signals (from/to gates), dotted lines are memory cell lanes used upon selection, dashed lines represent the carry lanes used otherwise; dotted and dashed lanes are mutally exclusive}
    \label{fig:attLSTM}
\end{figure}

We allow hidden state to $choose$ a memory cell that it $wants$ to $read$ from and $write$ to. This choice will be invisible to other hidden states, so that multiple memory cell choices can be mapped to the same internal state, giving some space to learning invariant temporal patterns. Compared to the original LSTM and simple Array-LSTM, one additional gate activation per memory lane is computed. We call it $selection$ gate. See Fig. \ref{fig:attLSTM} and equations (4.1-4.8) for details. The intuition behind this approach is that the $s$ gates control how much a particular memory lane should be used during current time step. If $s^t$ is 1 then all other activations remain unchanged, we fully $process$ this memory cell. If $s^t$ is 0, then the contents are carried over from the previous time step and the memory cell is not affected during that time step (no read/no write). The idea is that such a mechanism should allow less leaky memory cells, effectively not relying entirely on forget gate in order to preserve its contents. Each memory lane has a selection gate $s^t$ associated with it, controlling information flow through or bypassing current time step (carrying over memory content from previous time step). The idea is that $s$ gates should be responsible for controlling the timescale (as proposed in the Zoneout paper \citep{DBLP:journals/corr/KruegerMKPBKGBL16}). \\

Attention signals $k$.
\begin{equation}
a_k^t = \sigma({W_{ak} x^t + U_{ak} h^{t-1} + b_{ak}})
\end{equation}

Softmax normalization.

\begin{equation}
s_k^t = \frac{e^{a^t_k}}{\sum_{k}{e^{a^t_k}}}
\end{equation}

\begin{equation}
f_k^t = s_k^t \odot \sigma({W_{fk} x^t + U_{fk} h^{t-1} + b_{fk}})
\end{equation}
\begin{equation}
i_k^t =  s_k^t \odot \sigma({W_{ik} x^t + U_{ik} h^{t-1} + b_{ik}})
\end{equation}
\begin{equation}
o_k^t =  s_k^t \odot \sigma({W_{ok} x^t + U_{ok} h^{t-1} + b_{ok}})
\end{equation}

$\tilde{c}_k^{t}$ is not affected by $s_k^t$.

\begin{equation}
\tilde{c}_k^{t} = tanh({W_{ck} x^t + U_{ck} h^{t-1} + b_{ck}})
\end{equation}

Note the inverted $f$ gate effect for clearer notation (cell $k$ is reset for $f_k$ = 1). If $s_k^t$ is 0, then $f_k^t$ is 0 and effectively $k$ memory cell's contents are entirely transferred for the previous time step.
\begin{equation}
c_k^{t} = (1-f_k^t) \odot c_k^{t-1} + i_k^t \odot \tilde{c}_k^{t}
\end{equation}
\begin{equation}
h^{t} = \sum_{k}{o_k^t} \odot tanh(c_k^{t})
\end{equation}

\paragraph{Max pooling version}
\label{maxsection}
In this version the algorithm used a hard, deterministic selection, by choosing the lane with highest $s^t$ value. It ignores other lanes and  backpropagates errors only through that lane, It is analogous to max pooling in CNNs.

\section{Non-deterministic Array-LSTM extensions}

\subsection{Stochastic Output Pooling}
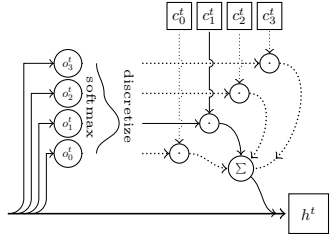
\begin{figure}[h]
 \centering 
 \begin{tikzpicture}[]
    \foreach \y [evaluate = \y as \expr using (\y*0.4-1.2), evaluate = \y as \exprr using (\y-3.5), evaluate = \y as \exprrr using  int(3-\y)]  in {0,1,2,3} {
   \node[draw,circle, xscale=0.45, yscale=0.45] at (6.7,\expr-4) (o_block\y) {$o_\y^{t}$};
   \node[draw,circle, xscale=0.6, yscale=0.6] at (\expr+9.38,\expr-4) (oc_block\y) {.};
   \node[draw,xscale=0.6, yscale=0.6]at (\expr+9.38,-3.37) (input\y) {$c_{\y}^{t}$};
   
    }

    \node[block, xscale=0.6, yscale=0.6] at (9.9,-6) (h2) {$h^{t}$};

    \draw[->, line width=0.6] (h2) ++ (-4,0) -- (h2);

    \node[draw,circle, xscale=0.4, yscale=0.4]at (9,-5.4) (sum) {$ \sum $};

    \draw[<-, line width=0.3] (o_block0.west) ++(-0.0,0) -- ++(-0.1,0) -- ++(0,-0.6) to [out=-90,in=0] ++(-0.2,-0.2);
    \draw[<-, line width=0.3] (o_block1.west) ++(-0.0,0) -- ++(-0.2,0) -- ++(0,-1.0) to [out=-90,in=0] ++(-0.2,-0.2);
    \draw[<-, line width=0.3] (o_block2.west) ++(-0.0,0) -- ++(-0.3,0) -- ++(0,-1.4) to [out=-90,in=0] ++(-0.2,-0.2);
    \draw[<-, line width=0.3] (o_block3.west) ++(-0.0,0) -- ++(-0.4,0) -- ++(0,-1.8) to [out=-90,in=0] ++(-0.2,-0.2);

    \draw[densely dotted, <-, line width=0.3] (oc_block0) -- ++(-0,0.35) arc(-90:90:0.05cm) -- ++(-0,0.3) arc(-90:90:0.05cm) -- ++(-0,0.3) arc(-90:90:0.05cm) -- ++(-0,0.37);
    \draw[<-, line width=0.3] (oc_block1) -- ++(-0,0.35) arc(-90:90:0.05cm) -- ++(-0,0.3) arc(-90:90:0.05cm) -- ++(-0,0.37);
    \draw[densely dotted, <-, line width=0.3] (oc_block2) -- ++(-0,0.35) arc(-90:90:0.05cm) -- ++(-0,0.37);
    \draw[densely dotted,<-, line width=0.3] (oc_block3) -- ++(-0,0.43);

    \draw[line width=0.3] -- (oc_block3) ++(-2.3, -1.2) 
    to [out=90,in=-90] ++(0.3, 0.4)
    to [out=90,in=-90] ++(-0.2, 0.4)
    to [out=90,in=-90] ++(-0.1, 0.4);

    \draw[line width=0.3] (o_block0.east) -- (o_block0.east) ++(0.2,0);
    \draw[line width=0.3] (o_block1.east) -- (o_block1.east) ++(0.3,0);
    \draw[line width=0.3] (o_block2.east) -- (o_block2.east) ++(0.5,0);
    \draw[line width=0.3] (o_block3.east) -- (o_block3.east) ++(0.2,0);

    \draw[densely dotted, ->, line width=0.5] (o_block0.east) ++(0.8,0) -- (oc_block0.west);
    \draw[, ->, line width=0.3] (o_block1.east) ++(0.8,0) -- (oc_block1.west);
    \draw[densely dotted, ->, line width=0.5] (o_block2.east) ++(0.8,0) -- (oc_block2.west);
    \draw[densely dotted, ->, line width=0.5] (o_block3.east) ++(0.8,0) -- (oc_block3.west);

    \draw[densely dotted,->-=.90, line width=0.5] -- (oc_block3) to [out=0,in=0] (sum);
    \draw[densely dotted, ->-=.99, line width=0.5] -- (oc_block2) to [out=0,in=45] (sum);
    \draw[->-=.99, line width=0.3] -- (oc_block1) to [out=0,in=90] (sum);
    \draw[densely dotted,->-=.99, line width=0.5] -- (oc_block0) to [out=0,in=180] (sum);
    \draw[->-=.98, line width=0.3] -- (sum) to [out=-40,in=180] ++(0.5,-0.6);

\node[xscale=0.3, yscale=0.3, label={[label distance=0.5cm,text depth=-1ex,rotate=-90, font=\tiny]right:softmax}] at (7.05,-3.55) {};

\node[xscale=0.3, yscale=0.3, label={[label distance=0.5cm,text depth=-1ex,rotate=-90, font=\tiny]right:discretize}] at (7.55,-3.45) {};

\end{tikzpicture}
 \caption{Stochastic Output Pooling: Sample open output gate index $i$ from this distribution calculated using $softmax$ normalization. Dotted lines represent inactive connections (set to zero due to the sampling procedure).}
\label{fig:splstm}
\end{figure}

This is the simplest of the considered stochastic architectures (Fig. \ref{fig:splstm}). It works by treating initial $o$ gate activations as inputs to a softmax output distribution and sampling from this distibution. Therefore it enforces normalization of output response and sparse binary outputs ($\leqslant 1/2$). During backpropagation, the algorithm uses only the selected gate (as in the max attention algorithm in \ref{maxsection}). All other steps are exactly the same as the standard Array-LSTM approach in \ref{arraysection}.

Probability that memory cell $i$ will used during $h^t$ update (other cells' outputs are not used): 

\begin{equation}
p(i = k) = \frac{e^{o^t_k}}{\sum_{k}{e^{o^t_k}}}
\end{equation}
The hidden state $h^{t}$ is computed using the active output cell index.
\begin{equation}
h^{t} = o_i^t \odot tanh(c_i^{t})
\end{equation}
\subsection{Stochastic Memory Array}
\label{stocharray}
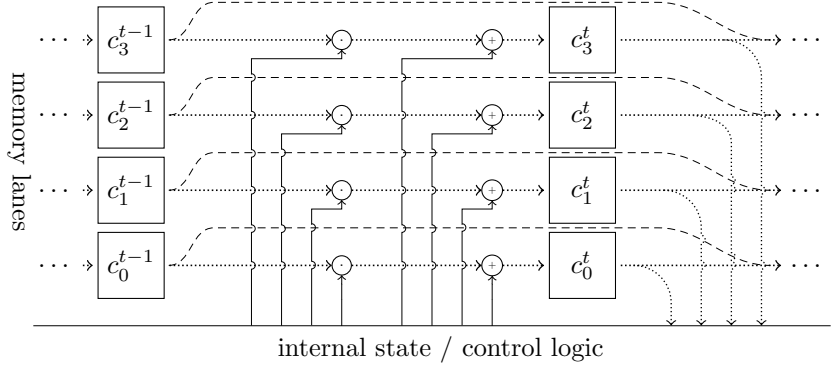
\begin{figure}
 \centering
 \begin{tikzpicture}
    \foreach \y [evaluate = \y as \expr using (\y*0.4-1.2), evaluate = \y as \exprr using (\y-3.5), evaluate = \y as \exprrr using  int(3-\y)]  in {0,1,2,3} {
           \node[] at (0,-\y) (input_prev\y) {$\dots$};
        \node[block] at (1,-\y) (input\y) {$c_{\exprrr}^{t-1}$};
        \node[draw,circle, xscale=0.6, yscale=0.6]at (3.8,-\y) (f_\y) {.};
         \node[draw,circle, xscale=0.4, yscale=0.4]at (5.8,-\y) (f2_\y) {+};

        \draw[<-, line width=0.3] (f_\y) -- ++(0,-.25) -- ++(\expr,0) -- ++(0,-.21);
        \draw[<-, line width=0.3 ] (f2_\y) -- ++(0,-.25) -- ++(\expr,0) -- ++(0,-.21);
        \node[block] at (7,-\y) (block\y) {$c_{\exprrr}^t$};
        \node[] at (10,-\y) (input_next\y) {$\dots$};
        \draw[densely dotted, ->, line width=0.6] (input_prev\y.east) -- (input\y);
        \draw[densely dotted, ->, line width=0.6] (input\y) --  (f_\y.west);
    \draw[densely dotted, ->, line width=0.6] (f_\y.east) --(f2_\y.west);
    \draw[densely dotted, ->, line width=0.6] (f2_\y.east) --(block\y);
        \draw[densely dotted, ->, line width=0.6] (block\y.east) -- (input_next\y);
        
    }

\draw[solid, line width=0.2]  -- ([xshift=-5cm, yshift=-10.82em]f2_0) ++(-1.1, 0) -- ++(10.6, 0);

\node[label={[label distance=0.5cm,text depth=-1ex,rotate=-90]right:memory lanes}] at (-0.5,0.2) {};
\node[label={[label distance=0.5cm,text depth=-1ex,rotate=0]right:internal state / control logic}] at (2.2,-4) {};

    \draw[line width=0.3]  -- ([xshift=-1.2cm, yshift=-1.85em]f_0) ++(0,.2)
    arc(90:-90:0.05cm) -- ++(0,-.15)
    -- ++(0,-.25) arc(90:-90:0.05cm) 
    -- ++(0,-.41) arc(90:-90:0.05cm)
    -- ++(0,-.39) arc(90:-90:0.05cm)
    -- ++(0,-.41) arc(90:-90:0.05cm)
    -- ++(0,-.39) arc(90:-90:0.05cm)
    -- ++(0,-.75);

    \draw[line width=0.3]  -- ([xshift=-0.8cm, yshift=-2.28em]f_0)  ++(0,-.65)
    arc(90:-90:0.05cm) -- ++(0,-.15)
    -- ++(0,-.25) arc(90:-90:0.05cm) 
    -- ++(0,-.41) arc(90:-90:0.05cm)
    -- ++(0,-.39) arc(90:-90:0.05cm)
    -- ++(0,-.75);

    \draw[line width=0.3]  -- ([xshift=-0.4cm, yshift=-5.82em]f_0) ++(0,-0.4)
    arc(90:-90:0.05cm) -- ++(0,-.15)
    -- ++(0,-.25) arc(90:-90:0.05cm) 
    -- ++(0,-0.75);

    \draw[line width=0.3]  -- ([xshift=-0.0cm, yshift=-5.82em]f_0) ++(0,-1.41) -- ++(0,-0.35);

    \draw[line width=0.3]  -- ([xshift=-1.2cm, yshift=-1.85em]f2_0) ++(0,.2)
    arc(90:-90:0.05cm) -- ++(0,-.15)
    -- ++(0,-.25) arc(90:-90:0.05cm) 
    -- ++(0,-.41) arc(90:-90:0.05cm)
    -- ++(0,-.39) arc(90:-90:0.05cm)
    -- ++(0,-.41) arc(90:-90:0.05cm)
    -- ++(0,-.39) arc(90:-90:0.05cm)
    -- ++(0,-.75);

    \draw[line width=0.3]  -- ([xshift=-0.8cm, yshift=-2.28em]f2_0)  ++(0,-.65)
    arc(90:-90:0.05cm) -- ++(0,-.15)
    -- ++(0,-.25) arc(90:-90:0.05cm) 
    -- ++(0,-.41) arc(90:-90:0.05cm)
    -- ++(0,-.39) arc(90:-90:0.05cm)
    -- ++(0,-.75);

    \draw[line width=0.3]  -- ([xshift=-0.4cm, yshift=-5.82em]f2_0) ++(0,-0.4)
    arc(90:-90:0.05cm) -- ++(0,-.15)
    -- ++(0,-.25) arc(90:-90:0.05cm) 
    -- ++(0,-0.75);

    \draw[line width=0.3]  -- ([xshift=-0.0cm, yshift=-5.82em]f2_0) ++(0,-1.41) -- ++(0,-0.35);

    \draw[densely dotted, <-, line width=0.5] (oc_block0) ++(-0,1.4) -- ++(-0,0.3) to [out=90,in=0] ++(-0.5,0.5);

    \draw[densely dotted, <-, line width=0.5] (oc_block1) ++(-0,1.0) -- ++(-0,0.75) arc(-90:90:0.05cm) -- ++(-0,0.25) arc(-90:90:0.05cm) -- ++(0,0.1) to [out=90,in=0] ++(-0.5,0.5);

    \draw[densely dotted, <-, line width=0.5] (oc_block2) ++(-0,0.6) -- ++(-0,0.75) arc(-90:90:0.05cm) -- ++(-0,0.02) arc(-90:90:0.05cm) -- ++(-0,0.79) arc(-90:90:0.05cm) -- ++(-0,0.02) arc(-90:90:0.05cm) -- ++(0,0.45) to [out=90,in=0] ++(-0.5,0.36);

    \draw[densely dotted, <-, line width=0.5] (oc_block3) ++(-0,0.2) -- ++(-0,0.75) arc(-90:90:0.05cm) -- ++(0,0.9) arc(-90:90:0.05cm) -- ++(0,0.9) arc(-90:90:0.05cm) -- ++(0,0.45) to [out=90,in=0] ++(-0.5,0.5);

    \draw[densely dashed, line width=0.3] (input0) ++(0.5,0) to [out=0,in=180] ++(0.6,0.5) -- ++(5.9,0) to [out=0,in=180] ++(1.5,-0.5);
    \draw[densely dashed, line width=0.3] (input1) ++(0.5,0) to [out=0,in=180] ++(0.6,0.5) -- ++(5.9,0) to [out=0,in=180] ++(1.5,-0.5);
    \draw[densely dashed, line width=0.3] (input2) ++(0.5,0) to [out=0,in=180] ++(0.6,0.5) -- ++(5.9,0) to [out=0,in=180] ++(1.5,-0.5);
    \draw[densely dashed, line width=0.3] (input3) ++(0.5,0) to [out=0,in=180] ++(0.6,0.5) -- ++(5.9,0) to [out=0,in=180] ++(1.5,-0.5);

    \end{tikzpicture}
    \caption{Stochastic Memory Array: solid lines are control logic signals (from/to gates), dotted lines are memory cell lanes used upon selection, dashed lines represent the carry lanes used otherwise; dotted and dashed lanes are mutally exclusive}
    \label{fig:doLSTM}
\end{figure}

This is a more complex architecture, an $active$ cell is being chosen for the entire cycle of computation. Instead of selecting only the output memory cell, the algorithm chooses one memory lane to be used during the entire forward and backward loop. If a lane is not selected, carry over cell contents. This idea is very similar to the algorithm described as Zoneout \citep{DBLP:journals/corr/KruegerMKPBKGBL16}. The main difference between two implementations is the fact that here, the hidden state is not affected directly by the noise injecting procedure. Instead, by choosing exactly one memory lane, it ensures that the output of the memory is non-zero and the content of $h^t$ is going to be non-zero (I found that the main instability of all stochastic approaches, including Recurrent Dropout \citep{DBLP:journals/corr/ZarembaSV14} is caused by the fact that all or almost all hidden states' activations can be zeros. Figure \ref{fig:doLSTM} depicts the changes to the original Array-LSTM architecture from \ref{arraysection}. Each memory lane has an additional bypass connection which is used when memory cell is not selected for computation (as in the soft attention algorithm in \ref{softatt}). We consider 2 variants of this approach: (a) 1 out of K cells is active (b) 1/2 cells are active, i.e. 4 out of 8. In the first implmentation, the algorithm randomly chooses index of the memory cell to be used. In the latter one, the same procedure applies, only to groups of 2 cells (odd or even).


\subsubsection{Stochastic hard attention}

Here this report describes different implementations of combinining the soft attention mechanism with stochastic lane selection. We would like to select memory cell in a semi-random way, according to some selection distribution as described in \ref{softatt}.

\paragraph{Semi-Hard} One memory lane is selected as being active (as in Stochastic Memory Array from \ref{stocharray}). However, unlike the previous fully random selection mechanism, in this approach the lane is selected according to softmax distribution where $s$ are inputs controlling this distribution. The backward step is the same as in the soft attention version, ignoring the fact that in the forward pass a sample was used and computes derivatives using differentiable softmax outputs.

\paragraph{Hard} Same as v1, but the backward pass uses only the selected lane (as in \ref{stocharray})

\subsection{Implementation considerations}
Array-LSTM brings many advantages from the computation cost perspective. 
\begin{enumerate}
\item \textbf{More memory cells} Given the same amount of memory for allocation, Array-LSTM has effectively use more memory cells due to smaller matrix between hidden units and gates.
\item \textbf{More parallelism} More cells give raise to more independent elementwise computation which is very cheap on SIMD hardware like GPUs.
\item \textbf{More data locality} Cells belonging to the same hidden unit will be somehow correlated which might improve cache hit ratio. In addition to that, sparse hidden to hidden connectivity might be possible for larger number of cells per unit.
\item \textbf{Approximate} Stochastic Array Memory is naturally resilient to noisy input.
\end{enumerate}

\section{Related Works}
There are many other approaches to improving baseline LSTM architecture including deterministic approaches such as Depth gated LSTM \citep{DBLP:journals/corr/YaoCVDD15}, Grid LSTM \citep{DBLP:journals/corr/KalchbrennerDG15} or Adaptive Computation Time \citep{DBLP:journals/corr/Graves16}. Examples of stochastic variants include recurrent dropout \citep{DBLP:journals/corr/ZarembaSV14, DBLP:journals/corr/SemeniutaSB16} Batch normalization \citep{DBLP:journals/corr/CooijmansBLC16} and Zoneout \citep{DBLP:journals/corr/KruegerMKPBKGBL16}.

\section{Experiments}

\subsection{Methodology}


The learning algorithm used was backprogagation through time (BPTT), it proceeded by selecting random sequences of length 10000 randomly from a given corpus. The learning algorithm used was Adagrad\footnote{with a modification taking into consideration only recent window of gradient updates} with a learning rate of 0.001. Weights were initialize using so-called Xavier initialization \cite{Glorot10understandingthe}. Sequence length for BPTT was 75 and batch size 128, states were carried over for the entire sequence of 10000 simulating full BPTT. Forget bias was set initially to 1. These values were not tuned, so it is possible that better results can be easily obtained with different settings. The algorithm was written in C++ and CUDA 8 and ran on GTX Titan GPU for up to 20 days. Link to the code is at the end.


\subsection{Data}

\subsubsection{enwik8}
It constitutes first $10^8$ bytes of English Wikipedia dump (with all extra symbols present in xml), also known as Hutter Prize challenge dataset\footnotemark[1].
\subsubsection{enwik9}
This dataset is used in Large Text Compression Benchmark - the first $10^9$ bytes of the English Wikipedia dump\footnotemark[1].
\subsubsection{enwik10}
This dataset is an extension of enwik9 dataset and was created by taking first $10^{10}$ bytes of english wikipedia dump from June 1, 2016\footnote[2]{https://www.dropbox.com/s/kzb5a0bih99ltui/enwik10.txt}, entire dump\footnote[3]{https://www.dropbox.com/s/1wigbsjpwxtwh2k/enwiki-20160601-pages-articles.xml} (57G).

\subsubsection{Test data}
First 90\% of each corpus was used for training, the next 5\% for validation and the last 5\% for reporting test accuracy.

\subsection{Results}

It it not known what the limit on the compression ratio is beforehand. It it not known if a pattern is compressible, before actually being able to compress it or proving that it cannot be compressed. It was shown that for humans the perceived prediction quality of natural english text depends largely on the amount of context given and the ability to incorporate previous knowledge into making predictions and ranges from 0.6 to 1.3 depending on the case \citep{shannon1951prediction}{}. Another estimate was obtained by bzip and cmix9 algorithms used for text compression which prove the upper limit on the compressibility of these datasets (Table \ref{tab:table2}).

\begin{table}[h!]
  \centering
  \caption{Bits per character (BPC) on the Hutter Wikipedia dataset (test data). The results were collected using networks of approximately equal size (66M parameters)}
  \label{tab:table2}
  \begin{tabular}{rccc}
    \toprule
    & enwik8 & enwik9 & enwik10\\
    \midrule
    bzip2  & 2.32 & 2.03 & - \\
    
    \midrule
    mRNN\footnotemark[1]\citep{ICML2011Sutskever_524} & 1.60 & 1.55 & - \\
    GF-RNN \citep{DBLP:journals/corr/ChungGCB15} &  1.58 & - & - \\
    Grid LSTM \citep{DBLP:journals/corr/KalchbrennerDG15} & 1.47 & - & - \\
    MI-LSTM \citep{DBLP:journals/corr/WuZZBS16} & 1.44 & -  & - \\
    Recurrent Highway Networks \citep{1607.03474} & 1.42 & -  & - \\
    \midrule
    LSTM (my impl)  & 1.45 & 1.13 & 1.24\\
    Stacked 2-LSTM (my impl)  & 1.468 & \textbf{1.12} & \textbf{1.19}\\
    Vanilla Array-4 LSTM  & 1.445 & - & -\\
    Soft Attention Array-2 LSTM   & 1.43 & - & -\\
    Hard Attention Array-2 LSTM   & 1.43 & - & -\\
    Output Pooling Array-2 LSTM & 1.422 & - & -\\
    Stochastic Lane Array-2 LSTM & \textbf{1.402} & - & - \\
    \bottomrule
    cmix9\footnotemark[2]  & 1.25 & 0.99 & -\\
  \end{tabular}
  
\end{table}
\footnotetext[1]{preprocessed data, with 86-symbol alphabet, 2G dataset instead of enwik9}
\footnotetext[2]{http://mattmahoney.net/dc/text.html}
%

\begin{figure}
 \centering 
 \includegraphics[scale=0.5]{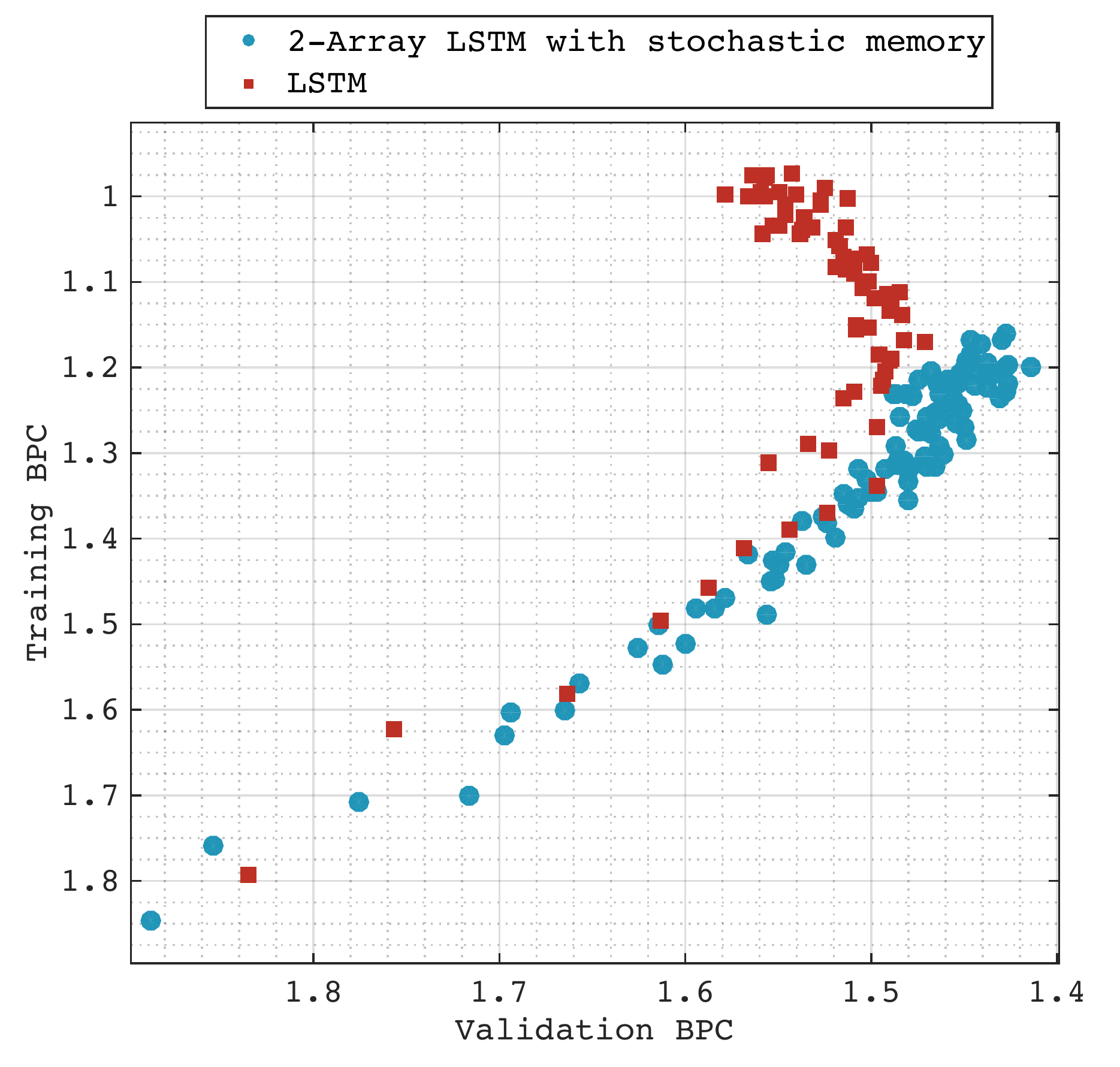}
 \caption{Training progress on enwik8 corpus (validation vs training), bits/character) - overfitting}
\label{fig:curves}
\end{figure}


\begin{figure}
 \centering 
 \includegraphics[scale=0.41]{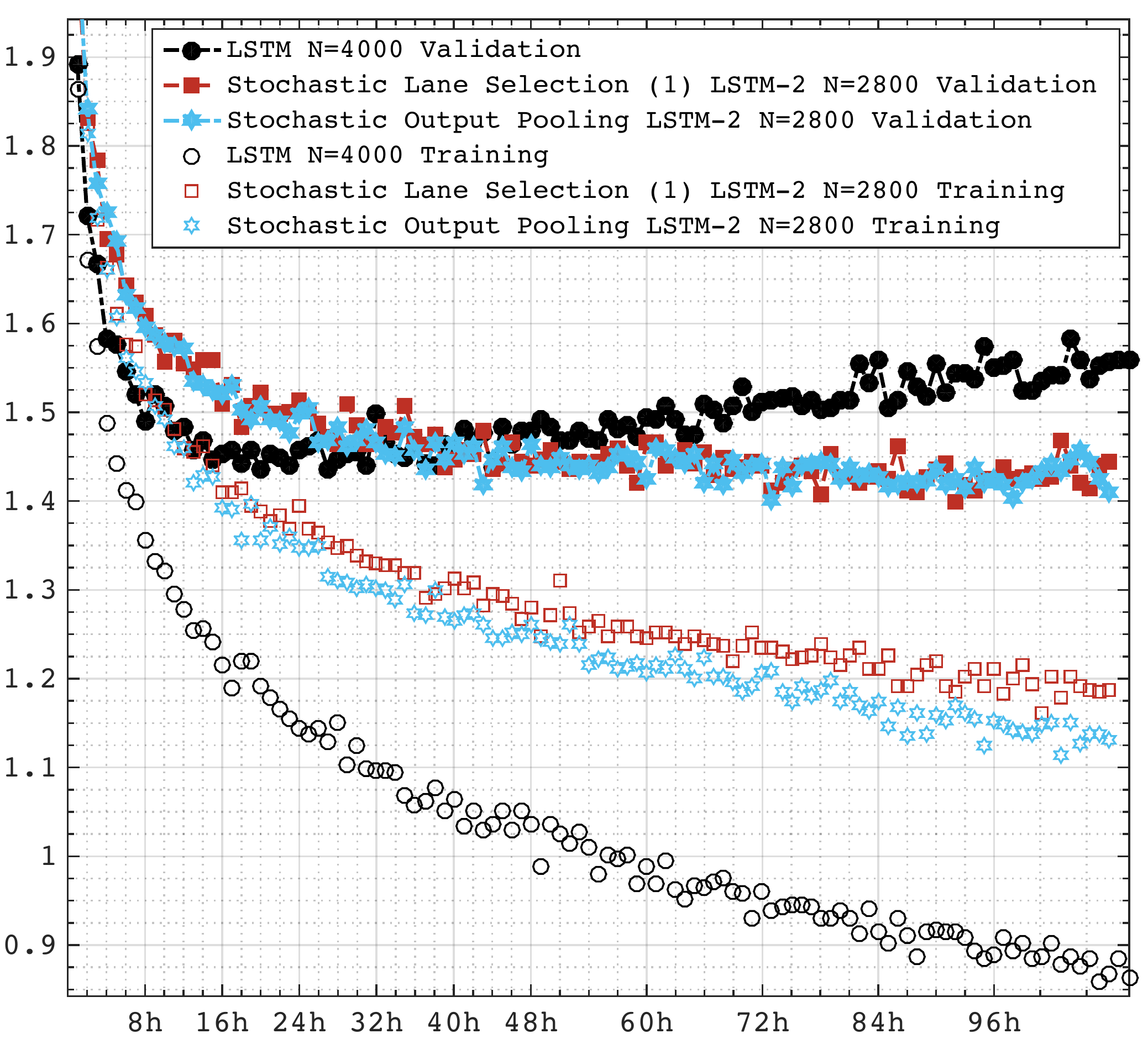}
 \caption{Training progress on enwik8 corpus, bits/character)}
\label{fig:stochastic_array}
\end{figure}




\begin{figure}[h]
 \centering 
 \includegraphics[scale=0.4]{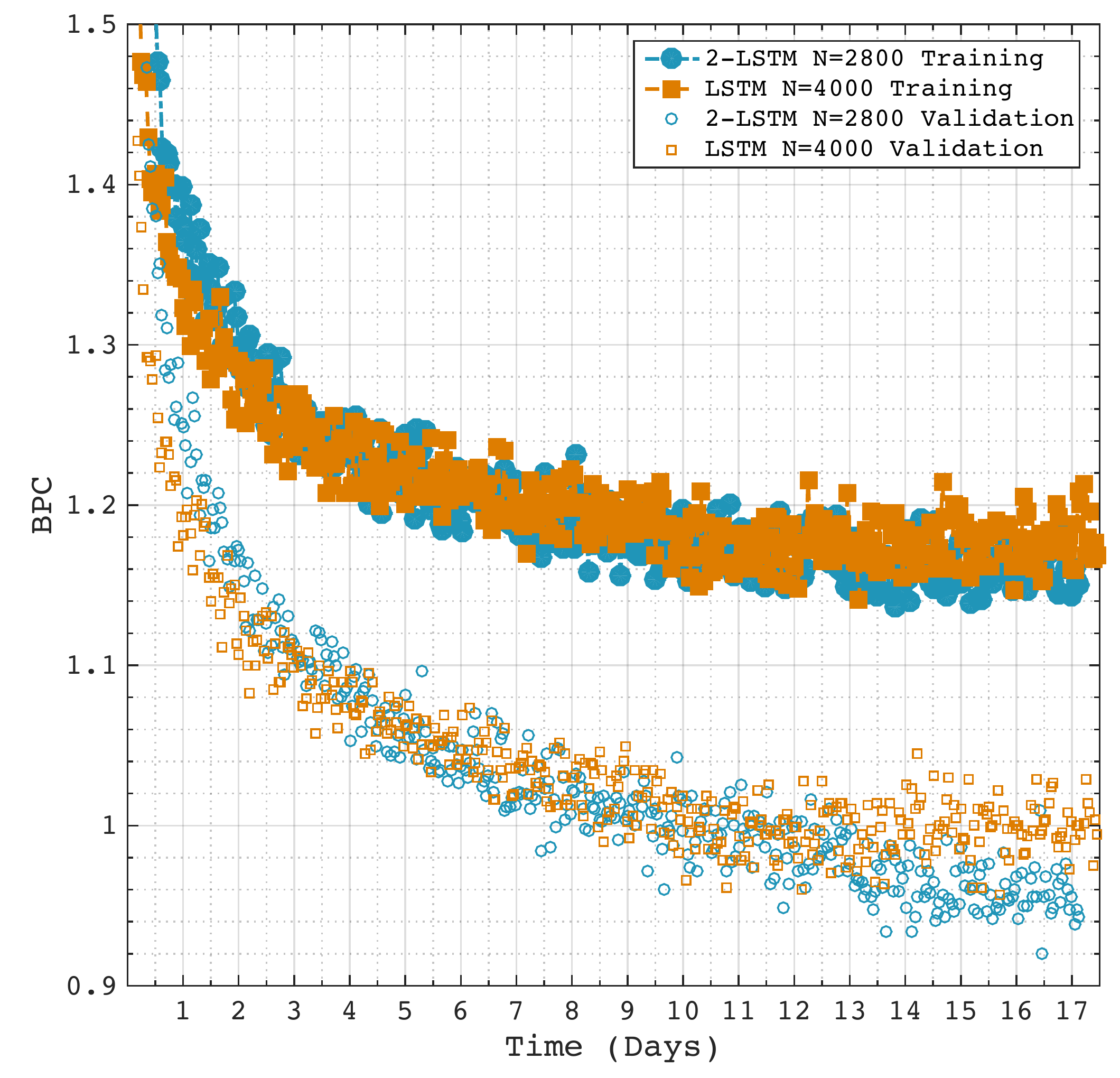}
 \caption{Training progress (x-axis is time) on enwik9 corpus. Problem seems to be capacity limited for this dataset, after 2 weeks of learning both the training loss and validation loss are still decreasing. Learning proceeds in a very similar way on enwik10 dataset (both curves are slightly higher, approximately 0.05 bits). Array-LSTM architectures did not provide any boost in learning convergence, comparable performance. }
\label{fig:enwik9}
\end{figure}

\subsection{Observations}

\subsubsection{Worked}
\begin{enumerate}
\item \textbf{Array-LSTM performs as well as regular LSTM}; it seems that sharing control states does not affect negatively training as long as there is the same number of memory cells), despite having limited number of hidden nodes (but same number of parameters). However, like LSTM it exhibits strong overfitting on enwik8 datasets (after about 24h of training). In some cases convergence speed is better when using state sharing, however it does not seem to generalize better than standard LSTM contrary to expectations.
\item \textbf{Stochastic Memory Array} - need to make sure that at least 1 cell is on, otherwise it blows up.
\end{enumerate}
\subsubsection{Inconclusive}
\begin{enumerate}
\item \textbf{Hard attention} - couldn't make it work well, converged slowly, hard to debug. Max variant works slightly better, but not results are not convincing.

\item \textbf{Attention modulated LSTM} comparable to Array-LSTM, not much of an improvement.
\item \textbf{Stochastic Output Pooling} works if $n=2$, for $n>2$ there is too much randomness (for $n=4$, effectively the dropout rate is 75 percent, training is slow), Stochastic Memory Array works better with 1/2 cells active.
\end{enumerate}
\subsubsection{Did not work}
\begin{enumerate}
\item Uncostrained Dropout (allowing all cells or states to be zeros - causes unstable behavior).
\item Multiplicative interactions between memory cells within one column/array or stack (example - \ref{fig:stack})

\end{enumerate}

\subsubsection{Other observations}
\begin{enumerate}
\item 2-LSTM or multilayer LSTM do not really improve things, no gain in generalization, slightly better capacity observed on enwik10; a better way of injecting compositionality is required. In short, one large layer is equally capable, which may be explained by inherent deep structure of recurrent nets.
\item No visible overfitting with standard LSTM on enwik9 and enwik10 after 2 weeks of training - 2 problems.
\item On enwik8, the main problem is overfitting, even small nets (less than 500 units) can basically memorize small corpora, observed that during generation of sequences it can $recite$ fragments of text, need to improve generalization - incorporate temporal invariance into architecture.
\item results from compression challenge (cmix9) show that there is some redundancy left, so it should be possible to get better results
\item Batch size affects convergence speed, sequence length (possibly because we carry last state over emulating full BPTT) and epoch length have low impact
\item Performance analysis supports persistent RNN paper \citep{diamos2016persistent}; Most time is spent moving data, pointwise OPs are very cheap
\item Qualitative evaluation Generated sequences give some hints, these are only hypotheses which need to be confirmed. cells may be sensitive to: \ldots
\begin{enumerate}
\item {position, dates, city names, languages, topics, numbers, quantities}
\end{enumerate}
An analysis as in \citep{DBLP:journals/corr/KarpathyJL15} may be required. See the Appendix for generated samples, we also provide nearly 5000 samples of length 5000 each as a by-product of the experiments for analysis.

\item Even BPC of 1.0 seems to be insufficient in order to generate human-level text, but getting close (heavily structured text is OK)
\end{enumerate}

\subsubsection{Further work}

\begin{enumerate}
\item How to incorporate compositionality in a better way to reflect recursive structure - feedback: how to implement it efficiently?
\item {How to do learning in a more efficient way which reflects the asynchronous nature of event.  BPTT works, but the time horizon is fixed, so how to make parts of network $see$ different past sequences, incorporate only relevant symbols, i.e. discarding exact timing and preserving ordering. }
\begin{enumerate}
\item \textbf{Stack vs Array, Tape} The data structure should enforce representation compositionality, we are experimenting with the following structures injecting feedback signal, but no significant improvement. We got it to work in practice, and there is some experimental evidence that it may increase capacity, but it took too much time on enwik9 and enwik10 datasets using current implementation to draw any definive conclusions.

\begin{figure*}[t!]
    \centering
    \begin{subfigure}[]{0.35\textwidth}
        \centering
        \includegraphics[height=2.7in]{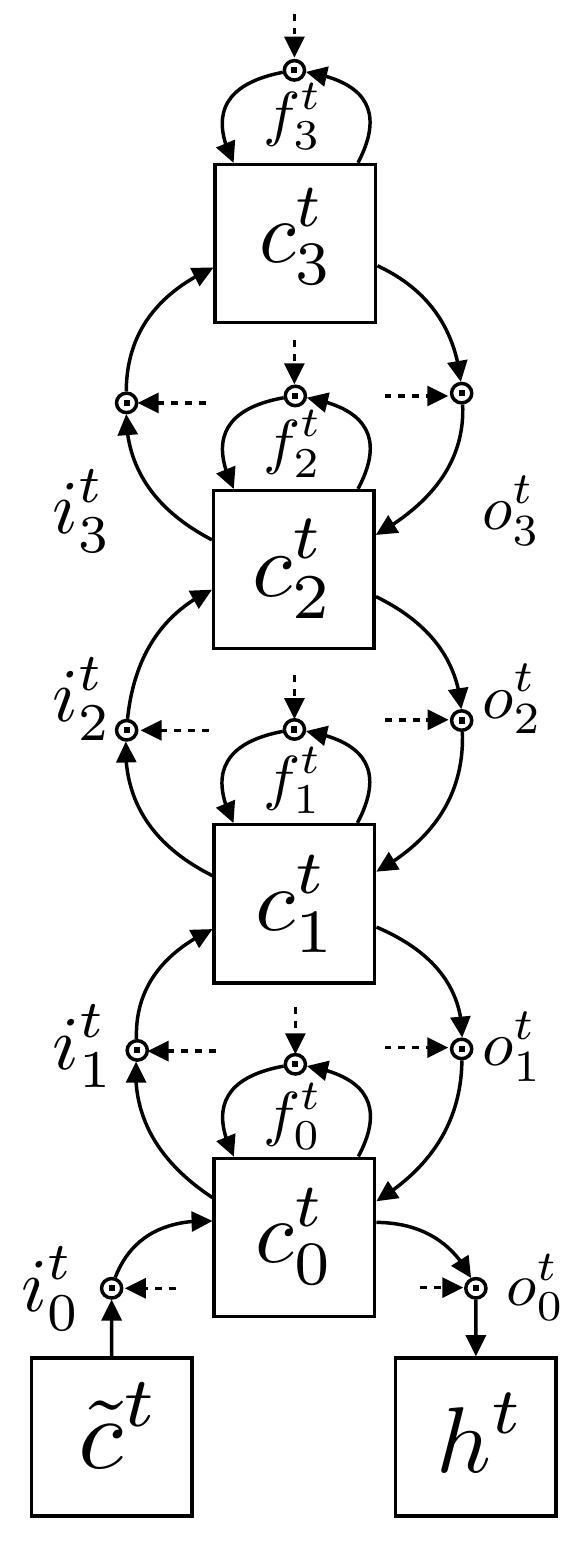}
        \caption{Stack-LSTM}
    \end{subfigure}%
    ~ 
    \begin{subfigure}[]{0.4\textwidth}
        \centering
        \includegraphics[height=2.5in]{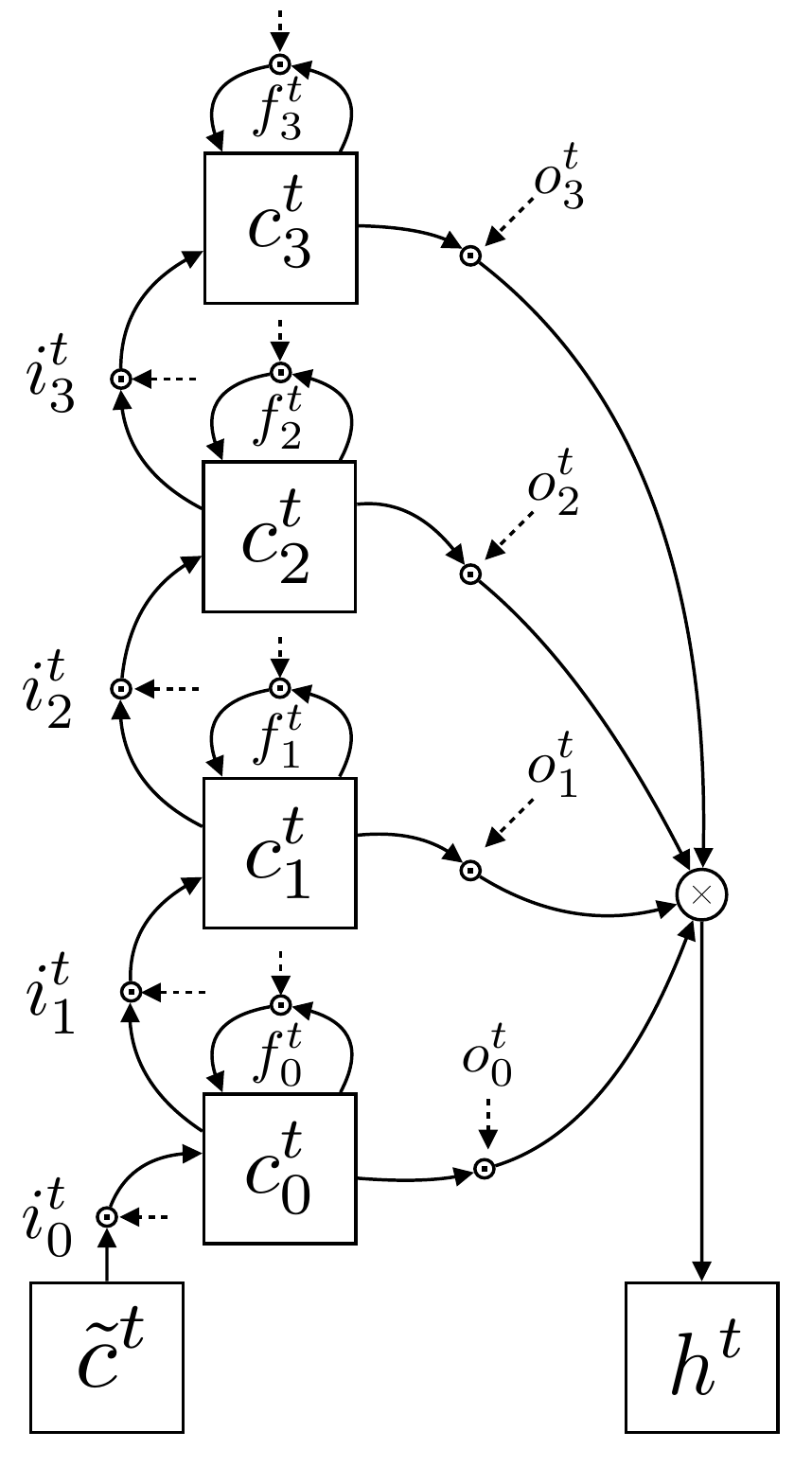}
        \caption{Stack-LSTM with Multiplicative Output connections}
    \end{subfigure}

    \caption{The main idea is that a stack-like structure enforces ordering in the data-flow between low and high frequency patterns; here we assume that bottom memory cells deal with high frequency and top cell with least frequent changes}
\label{fig:stack}
\end{figure*}

\begin{figure*}[t!]
    \centering
    \begin{subfigure}[]{0.6\textwidth}
        \centering
        \includegraphics[height=1.4in]{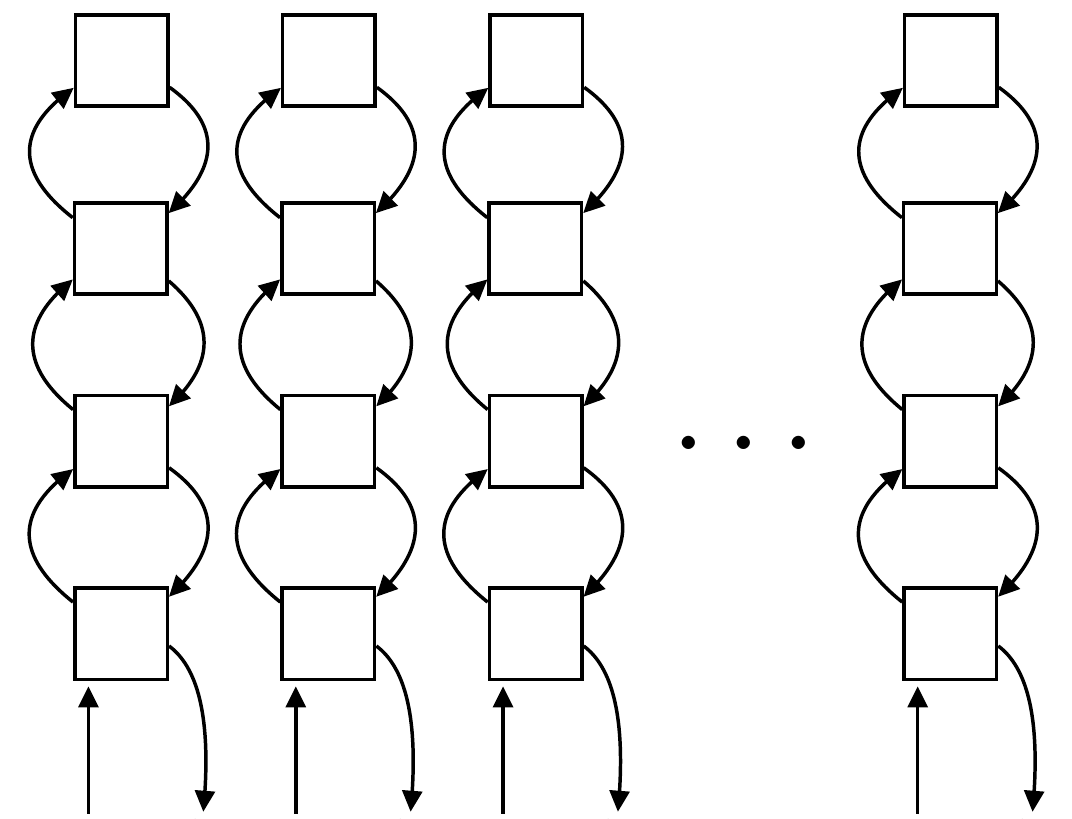}
        \caption{1-Layer Stack-LSTM with 4 cells/column}
    \end{subfigure}%
    ~ 
    \begin{subfigure}[]{0.4\textwidth}
        \centering
        \includegraphics[height=1.4in]{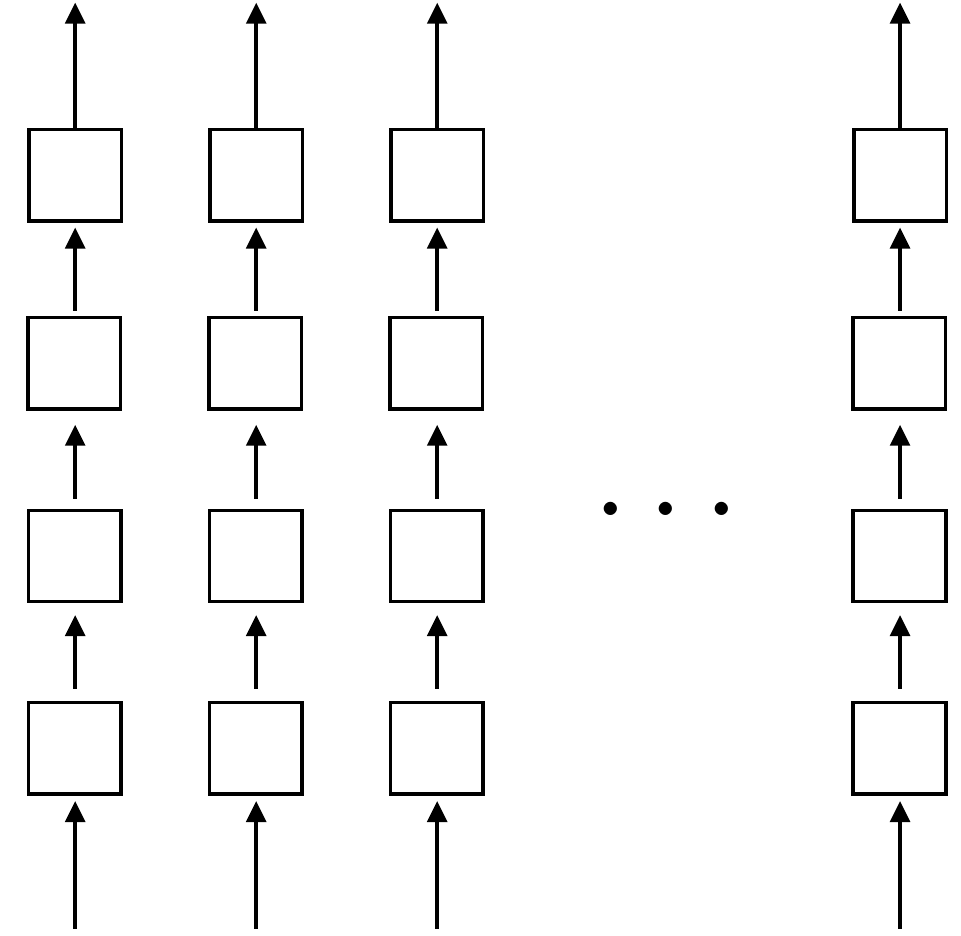}
        \caption{4-Layer Stacked LSTM}
    \end{subfigure}

    \caption{Stack-LSTM with the memory hierarchy built-in into the layer (1 layer with columnar cell stack) feedback architecture vs popular stacked-LSTM which is a feedward multilayer architecture without feedback from higher layers}
\label{fig:stack2}
\end{figure*}




\item {The relationship between gating-like mechanism in RNNs and thalamocortical circuits in the brain - resonant columns; there seem to be some clues that cortex incorporates some form of gating mechanism though thalamus, but we don't know exactly how it works and how feedback is implemented - some research is needed - see Appendix B}

\end{enumerate}
\item {What is the hard limit on BPC - will improving BPC lead to better samples? See Appendix A for low entropy samples from enwik9 and enwik10 datatests}

\end{enumerate}

\section{Conclusions}
This report presented Array-LSTM approaches. Stochastic memory operation seems to be necessary in order to mitigate overfitting effects. We tried many deterministic variants, but they all overfit as easily as baseline LSTM. Based on results from enwik9 and enwik10 datasets, the best regularization strategy is simply more data. Furthermore the Stochastic Memory Array approach extended state-of-the-art on enwik8 and set reference results for neural based algorithms on enwik9 and enwik10 datasets.

\section*{Acknoledgements}
This work has been supported in part by the Defense Advanced Research Projects Agency (DARPA).

\bibliography{alstm}

\begin{appendices}
\section{Generated samples} \label{App:AppendixA}
A gallery of samples generated by networks, standard procedure was used: initialize all $h$ and $c$ memory units with 0, generate one symbol at a time according to the output probability distribution conditioned on $h^t$, treat that input as the next input, iterate a few thousand times, see \url{https://github.com/krocki/ArrayLSTM/tree/master/samples} for more samples.

\begin{lstlisting}[caption=enwik8.txt Stochastic Array-2 LSTM N-4000 Validation BPC-1.395]
 well s lie to capture political activities created in and 2000 would be made up- if the Scripturals reached between ngation and decent from three to five, and when the painting followed, about 68,000 members of societies had not yet deployed its monumental calilets as I continued to come back towards them, liliced to familiarity. Most hinders thereore, concerning the behavioral science of the latter was done in the 1980s.
*Third participation did not imply that close relationships were annoyal because of the distinctive emotions, both intrinsic and realism.

(see [[Dark mass disaster]]). In particular, Darwinisms that arise out by meipsycological treatment because many is reearted in he response limit of inn. In some further cases, particularly apologetative sounds do not prefix conditions, such as results of non-coditional analogy in cles of gaining chemical in different ways will almost completely lead to death, or require some extent by emphasizing that in general times or death, types or etching.

This is not associated with [[Negative testiny]] or [[Gaut in the ulture]].

==Usage==
Within the [[Chinese experiment of sexual abuse]], '''evolution''' is also called &quot;[[Generalissimple kinese object]]&quot;.  It however is at least essentially understood and alleged by the philosophers of [[societal planning]] as a result of analogous working-nature expressions in overrun variables based on the expectations of which aid is inevitable.

==History of the name==
Criticism of fundamental differentialism by modern day liberties now refers to individual organizations such as [[Black paricular]]s. In [[Germany]], which examples as either a consonant or non-GFT, a correspondence against the people of the [[West End of Britain]]. Prior to the case was beginning in 1983 in relation to the subdivision of the cultural landforms of many years of academic life (at least once, according to 16th century chronology law   neither language 'ar' Esperanto had less so in [[Old Norse literature]] than on other balancing words. 

[[Old Scottish Gaelic language|Old English variations]] are be&quot;liked&quot; in [[Taggart Stanley]]'s &quot;New Yorkers.&quot;

*Romantic Cross-and-Sip Modern-Literature generators would retain him in [[Christmas]] notice as the composer of a child's chick in the Charles W. Burgh6 story in books or history; two L5 centuries.
The Dish Studios describe a giant as Ann She given in [[Weether University]]. This lingens the linement in the semi-final dayline: in his ack of main characters, Jeffrey was indeed his lyrical heart-town. However, the general rule would find confusing well, the film, the odds being exceeded. Another popular transgressionist [[Evan Tobi]], openly Wet artist, inspared Moore for the time. He also made many successes, including the sequel, ''[[Rolling Stone]]'', and a number of pseudoblasts. ''[[John Gapling Madden]]'' also starred [[Elizabeth Jones]] as ''Back and East''. 

Born in [[Scagblaw]], Buckingham Carolini were among the only serious visual concerns over one such theeto ad continued to exchange dance bolbloonfriends their involvement in devesong her being. A young Bruce Goldwater wife promoted William Randolly to I.L. in this year. In the end was Bell running back in Parliament with a bug in the [[odorn slug administration]].

During which three former top-ever, filed and bigenting factories in the upstate New York, factories family were typically severely defered; she sometimes refused.

Coupland also appeared in ''The Party'' and ''[[RuneQuest]]'' (''[[300 Begs For Studies]]'') in [[1968]]. It took the characters found at Berkeley in theater in [[Rochester, New York|Rochester]] on [[June 3]], [[1923]] at Anny Bryant London as an staple, after the [[[1904 in film|1945]] interview that the [[film|movie]] played the play.

After this peregrension, Beatle came from a group of [[Rock and Roll Hall of Fame]]rs. Much of his life is considered cross-time and less extreme and successful in popular drink. Her novel ''Places in Amoryo music'', penned for four plays selling the opening band [[Kazuma Ahmet]] (''[[Amalghend]]'') rolled out the [[United Kingdom|British]] movie of the 20th century and is most naïme out of the role.

==External links==
{{wikibooks}}
* {{imdb name|id=0416003|name=Charles Lindberg}}
* [http:/www.farleyfnet.com/ Far Easter.com] (unofficial) Quote
* [http://www.fantagnalsilot.co.uk Frankie Nobina, Andy Warhol]
* [http://www.friendsfootballsholdon.com/shaganshish.html Fall for Anarchy Fancifully Charted from The Golden Dawn of the 1913 General Theater]
* [http://www.cwvtc.us/archives/hackman/050802.pdf ''And there Were Today's ' Portrait of Andrew Jackson''] from a musical account of written work on Camus ]
* {{cite book | first=Francis | last= Catted | author=Astounding Spooks | other= | others= | publisher=Freedman | year=1986 | id=ISBN 1-854889-04-7}}

''G&amp;SCollingd:
* St. Claire Clarke, &quot;Casualties of Eden Young Check in Closet&quot; in ''W
\end{lstlisting}

\begin{lstlisting}[caption=enwik9.txt LSTM N-3200 BPC-1.116]
 spread allith the age of 18 year. The [[population density]] is 218.1/km&amp;sup2; (560.5/mi&amp;sup2;).  There are 3,298 housing units at an average density of 153.5/km&amp;sup2; (390.7/mi&amp;sup2;).  The racial makeup of the town is 98.62% [[White (U.S. Census)|White]], 0.07% [[African American (U.S. Census)|African American]], 0.13% [[Native American (U.S. Census)|Native American]], 0.11% [[Asian (U.S. Census)|Asian]], 0.02% [[Pacific Islander (U.S. Census)|Pacific Islander]], 0.09% from [[Race (U.S. Census)|other races]], and 0.48% from two or more races.  0.16% of the population are [[Hispanic (U.S. Census)|Hispanic]] or [[Latino (U.S. Census)|Latino]] of any race.

There are 2,000 households out of which 45.5% have children under the age of 18 living with them, 85.6% are [[Marriage|married couples]] living together, 3.2% have a female householder with no husband present, and 14.9% are non-families. 12.9% of all households are made up of individuals and 3.8% have someone living alone who is 65 years of age or older.  The average household size is 3.02 and the average family size is 3.38.

In the town the population is spread out with 26.2% under the age of 18, 8.0% from 18 to 24, 29.8% from 25 to 44, 22.4% from 45 to 64, and 12.7% who are 65 years of age or older.  The median age is 36 years.  For every 100 females there are 100.8 males.  For every 100 females age 18 and over, there are 101.5 males.

The median income for a household in the township is $35,621, and the median income for a family is $35,729. Males have a median income of $29,463 versus $16,071 for females. The [[per capita income]] for the township is $12,675.  3.4% of the population and 2.4% of families are below the [[poverty line]].  Out of the total population, 1.1% of those under the age of 18 and 3.9% of those 65 and older are living below the poverty line.

== External links ==
{{Mapit-US-cityscale|41.830224|-92.713274}}

[[Category:Villages in Illinois]]
[[Category:Fayette County, Illinois|&quot;Grand Isle,&quot;]]</text>
    </revision>
  </page>
  <page>
    <title>Rethshawn, Illinois</title>
    <id>111452</id>
    <vision>
      <id>16001292</id>
      <timestamp>2004-12-21T22:31:38Z</timestamp>
      <contributor>
        <username>Rambot</username>
        <id>6120</id>
      </contributor>
      <comment>Updated internal [[Geographic References|references]]. Added [[Template:Mapit-US-cityscale]] external links. Reorganized and cleaned up the page.</comment>
      <text xml:space="preserve">'''Otho''' is a town located in [[Lyon County, North Carolina]].  As of the [[2000]] census, the town had a total population of 380.

== Geography ==
[[Image:NCMap-doton-Saxonville.PNG|right|Location of Saxonville, North Carolina]]
Stockyard is located at 35&amp;deg;20'20&quot; North, 81&amp;deg;36'1&quot; West (35.543920, -81.635405){{GR|1}}.

According to the [[United States Census Bureau]], the town has a total area of 7.3 [[square kilometer|km&amp;sup2;]] (2.8 [[square mile|mi&amp;sup2;]]).  7.3 km&amp;sup2; (2.8 mi&amp;sup2;) of it is land and none of it is covered by water.

== Demographics ==
As of the [[census]]{{GR|2}} of [[2000]], there are 9,851 people, 4,251 households, and 2,736 families residing in the town.  The [[population density]] is 233.0/km&amp;sup2; (607.3/mi&amp;sup2;).  There are 3,102 housing units at an average density of 226.4/km&amp;sup2; (583.2/mi&amp;sup2;).  The racial makeup of the city is 86.82% [[White (U.S. Census)|White]], 5.93% [[African American (U.S. Census)|African American]], 0.04% [[Native American (U.S. Census)|Native American]], 15.67% [[Asian (U.S. Census)|Asian]], 0.07% [[Pacific Islander (U.S. Census)|Pacific Islander]], 1.04% from [[Race (U.S. Census)|other races]], and 1.64% from two or more races.  7.17% of the population are [[Hispanic (U.S. Census)|Hispanic]] or [[Latino (U.S. Census)|Latino]] of any race.

There are 408 households out of which 26.8% have children under the age of 18 living with them, 52.2% are [[Marriage|married couples]] living together, 8.5% have a female householder with no husband present, and 35.2% are non-families. 31.0% of all households are made up of individuals and 14.0% have someone living alone who is 65 years of age or older.  The average household size is 2.43 and the average family size is 3.03.

In the town the population is spread out with 25.6% under the age of 18, 6.0% from 18 to 24, 26.8% from 25 to 44, 24.9% from 45 to 64, and 14.9% who are 65 years of age or older.  The median age is 39 years.  For every 100 females there are 95.3 males.  For every 100 females age 18 and over, there are 91.4 males.

The median income for a household in the city is $25,708, and the median income for a family is $31,398. Males have a median income of $29,589 versus $16,750 for females. The [[per capita income]] for the town is $13,053.  22.3% of the population and 18.0% of families are below the [[poverty line]].  Out of the total population, 37.3% of those unde
\end{lstlisting}
\begin{lstlisting}[caption=enwik10.txt 2-LSTM N-2304 BPC-1.27]
         = 
|image/Parkinmapsfrom1IMWebTourism.UH-CIPTIFICAIN.jpg
|imagesize                = 240
|image_caption            = View of the Republic of Pitza Park in Nottingham.
|image_flag               = 
|image_seal               = 
|image_map               = NSW Map of Indiana cities locations in Indiana.svg
|map_caption              = Location in Indiana
|image_map1               = 
|mapsize1                 = 
|map_caption1             = 
|subdivision_type         = [[List of countries|Country]]
|subdivision_name         = [[United States]]
|subdivision_type1        = [[Political divisions of the United States|State]]
|subdivision_name1        = [[Indiana]]
|subdivision_type2        = [[List of counties in Indiana|County]]
|subdivision_name2        = [[Mississippi County, Indiana|Mississippi]]
|subdivision_type3        = [[List of townships in Indiana|Township]]
|subdivision_name3        = [[Holden Township, Coshocton County, Indiana|Holden]]
&lt;!-- Politics -----------------&gt;
|government_footnotes     =
|government_type          =
|leader_title             =
|leader_name              =
|leader_title1            =
|leader_name1             =
|established_title        =
|established_date         =

&lt;!-gme1901jphotemft.&lt;ref name=Mw=102&gt;[[Hibernian|''Michigan Historical Markers'']]. [http://www.ks.gov/facts/hebstr/geog/grossmap.htm Former Government History database]&lt;/ref&gt;
|established_title3      = Town
|established_date3        = 1863
|area_magnitude           =
|area_total_km2           = 1.48
|area_total_sq_mi         = 0.47
|area_land_sq_mi          = 0.17
|area_water_sq_mi         = 0.02
|area_water_percent       = 0
|area_urban_sq_mi         = 
|area_metro_sq_mi         = 
|area_blank1_title        = 
|area_blank1_title        = 
|area_blank1_km2          = 
|area_blank1_sq_mi        = 
&lt;!-- Population   -----------------------&gt;
|population_as_of         = [[2010 United States Census|2010]]
|population_est           = 9211
|pop_est_as_of            = 2012&lt;ref name=&quot;2012 Pop Estimate&quot;&gt;{{cite web|title=Population Estimates|url=http://www.census.gov/popest/data/cities/totals/2012/SUB-EST2012.html|publisher=[[United States Census Bureau]]|accessdate=2013-05-29}}&lt;/ref&gt;
|population_note          = 
|population_total         = 118
|population_density_km2   = 662.1
|population_density_sq_mi = 1832.3
|population_metro        = 
|population_density_metro_km2     = 
|population_density_metro_sq_mi   = 
|population_urban                 = 
|population_density_urban_km2     = 
|population_density_urban_sq_mi   = 
|population_blank1_title          = 
|population_blank1                = 
|population_density_blank1_km2    = 
|population_density_blank1_sq_mi  = 
&lt;!-- General information  ---------------&gt;
|timezone                 = [[North American Central Time Zone|Central (CST)]]
|utc_offset               = -6
|timezone_DST             = CDT
|utc_offset_DST           = -5
|area_land_km2            = 1.38
|area_water_km2           = 0
|area_footnotes           = &lt;ref name=&quot;census-g001&quot;/&gt;
|area_total_km2           = 1.23
|area_total_sq_mi         = 0.60
|area_land_sq_mi          = 0.59
|area_water_sq_mi         = 0

&lt;!-- Population --&gt;
|population_as_of         = [[2010 United States Census|2010]]
|population_est           = 655
|pop_est_as_of            = 2012&lt;ref name=&quot;2012 Pop Estimate&quot;&gt;{{cite web|title=Population Estimates|url=http://www.census.gov/popest/data/cities/totals/2012/SUB-EST2012.html|publisher=[[United States Census Bureau]]|accessdate=2013-05-28}}&lt;/ref&gt;
|population_footnotes     = &lt;ref name =&quot;FactFinder&quot;/&gt;
|population_total         = 364
|population_density_km2   = 650.2
|population_density_sq_mi = 1840.0

&lt;!-- General information --&gt;
|timezone                 = [[North American Central Time Zone|Central (CST)]]
|utc_offset              = -6
|timezone_DST             = CDT
|utc_offset_DST           = -5
|elevation_footnotes      = 
|elevation_m              = 472
|elevation_ft             = 1546
|coordinates_display      = inline,title
|coordinates_type         = region:US_type:city
|latd = 49 |latm = 18 |lats = 45 |latNS = N
|longd = 101 |longm = 21 |longs = 33 |longEW = W

&lt;!-- Area/postal codes &amp; others --&gt;
|postal_code_type         = [[ZIP code]]
|postal_code              = 58023
|area_code                = [[Area code 508|508]]
|blank_name               = [[Federal Information Processing Standard|FIPS code]]
|blank_info               = 48-15224&lt;ref name=&quot;GR2&quot;&gt;{{cite web|url=http://factfinder2.census.gov|publisher=[[United States Census Bureau]]|accessdate=2008-01-31|title=American FactFinder}}&lt;/ref&gt;
|blank1_name              = [[Geographic Names Information System|GNIS]] feature ID
|blank1_info              = 0475524 &lt;ref name=&quot;GR3&quot;&gt;{{cite web|url=http://geonames.usgs.gov|accessdate=2008-01-31|title=US Board on Geographic Names|p
\end{lstlisting}

\end{appendices}

\end{document}